%% file: main.tex
\documentclass{article} 
\usepackage{iclr2026_conference,times}

\input{math_commands.tex}

\usepackage{hyperref}
\usepackage{url}
\usepackage{times}
\usepackage{graphicx}
\usepackage{mathtools}
\usepackage{amsthm}
\usepackage{float}
\usepackage{algorithmic}
\usepackage{multirow}
\usepackage{amsmath,amssymb,amsfonts}
\usepackage{booktabs}
\usepackage{subcaption}
\usepackage{bm}
\usepackage{tikz}
\usetikzlibrary{shapes.geometric,arrows.meta}

\title{Grouping Nodes with known Value Differences: A lossless UCT-based Abstraction Algorithm}

\author{Robin Schmöcker \\
Institute for Information Processing\\
Leibniz University Hannover\\
Hannover, Germany \\
\texttt{schmoecker@tnt.uni-hannover.de} \\
\And
Alexander Dockhorn \\
SDU Metaverse Lab \\
University of Southern Denmark \\
Odense, Denmark \\
\texttt{adoc@mmmi.sdu.dk} \\
\And
Bodo Rosenhahn \\
Institute for Information Processing\\
Leibniz University Hannover\\
Hannover, Germany \\
\texttt{rosenhahn@tnt.uni-hannover.de} \\
}

\iclrfinalcopy 
\begin{document}

\maketitle

\begin{abstract}
 \input{sections/abstract}
\end{abstract}

\section{Introduction}
\label{sec:intro}
\input{sections/intro.tex}


\section{Foundations of automatic abstractions}
\label{sec:foundations}
\input{sections/foundations}

\section{Method}
\label{sec:method}
\input{sections/method/motivation}
\input{sections/method/kvda_framework}

\input{sections/method/implementation}

\section{Experiment setup}
\label{sec:experiment_setup}
\input{sections/experiment_setup}

\section{Experiments}
\label{sec:experiments}
\input{sections/experiments/experiments}

\section{Conclusion, Limitations and Future Work}
\label{sec:future_work}
\input{sections/sum_concl_fw}

\newpage 


\bibliography{references}
\bibliographystyle{iclr2026_conference}

\newpage
\appendix
\section{Supplementary Materials}
\label{sec:appendix}
\input{sections/appendix}

\end{document}

%% file: math_commands.tex
\usepackage{amsmath,amsfonts,bm}

\def\eqref#1{equation~\ref{#1}}

\def\1{\bm{1}}

\DeclareMathAlphabet{\mathsfit}{\encodingdefault}{\sfdefault}{m}{sl}
\SetMathAlphabet{\mathsfit}{bold}{\encodingdefault}{\sfdefault}{bx}{n}



%% file: sections/abstract.tex
A core challenge of Monte Carlo Tree Search (MCTS) is its sample efficiency, which can be improved by grouping state-action pairs and using their aggregate statistics instead of single-node statistics. On the Go Abstractions in Upper Confidence bounds applied to Trees (OGA-UCT) is the state-of-the-art MCTS abstraction algorithm for deterministic environments that builds its abstraction using the Abstractions of State-Action Pairs (ASAP) framework, which aims to detect states and state-action pairs with the same value under optimal play by analysing the search graph. ASAP, however, requires two state-action pairs to have the same immediate reward, which is a rigid condition that limits the number of abstractions that can be found and thereby the sample efficiency. In this paper, we break with the paradigm of grouping value-equivalent states or state-action pairs and instead group states and state-action pairs with possibly different values as long as the difference between their values can be inferred. We call this abstraction framework Known Value Difference Abstractions (KVDA), which infers the value differences by analysis of the immediate rewards and modifies OGA-UCT to use this framework instead. The modification is called KVDA-UCT, which detects significantly more abstractions than OGA-UCT, introduces no additional parameter, and outperforms OGA-UCT on a variety of deterministic environments and parameter settings.

%% file: sections/intro.tex
Research into non-learning-based decision-making algorithms such as Monte Carlo Tree Search (MCTS) \citep{BrownePWLCRTPSC12,KocsisS06} is an active field. On the one hand MCTS can be used for applications where a general on-the-fly applicable decision-making algorithm is needed such as Game Studios which rarely use Machine Learning (ML) based AI as they would have to be retrained whenever the game rules are modified (e.g. during development or patches). And on the other hand, though not the scope of this paper, foundational work in MCTS might potentially translate to improvements of ML algorithms such as Alpha Zero \citep{alphazero} that are built on MCTS. 

One way to improve MCTS is to reduce the search space by grouping states and actions in the current MCTS search tree to enable an intra-layer information flow \citep{uctJiang,AnandGMS15,OGAUCT} by averaging the visits and returns of all abstract action nodes in the same abstract node used for the Upper Confidence Bounds (UCB) formula in the tree policy, which increases the sample efficiency. 
One key strength of one of the state-of-the-art abstraction algorithms On the Go Abstractions in Upper Confidence bounds applied to Trees (OGA-UCT) \citep{OGAUCT} is its exactness in the sense that if OGA-UCT groups two state-action pairs in a search tree where all possible successors of each state-action pair have been sampled, only state-action pairs are grouped that have the same $Q^*$ value, i.e., they have the same value under subsequent optimal play. This exactness condition, however, comes at the cost that state or state-action pairs that only differ slightly in their $Q^*$ value cannot be detected. This issue was slightly alleviated with the introduction of $(\varepsilon_{\text{a}},\varepsilon_{\text{t}})$-OGA \citep{ogacad}, which is equivalent to $(\varepsilon_{\text{a}},0)$-OGA in deterministic environments that allows for small errors in the immediate reward or transition function when building an abstraction. However, this also brings two downsides with it. Firstly, it introduces two parameters, which makes tuning harder, and secondly, grouping non-value equivalent state-action pairs might even be harmful to the performance, as they can make convergence to the optimal action impossible.

\begin{figure}[h]
  \centering
  \includegraphics[width=0.3\linewidth]{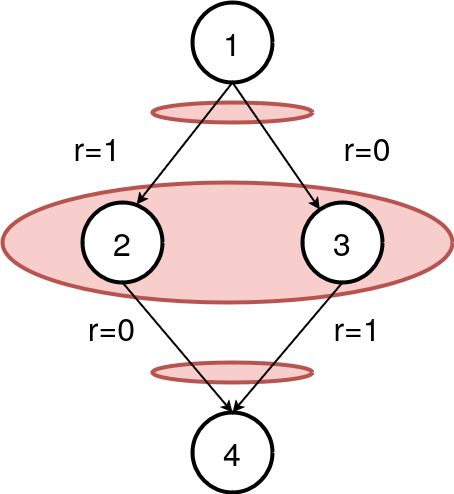}
  \caption{
    An example of an MDP state-transition graph where the state-of-the-art abstraction framework ASAP \citep{AnandGMS15} would detect no abstractions while \textbf{our method Known-Value-Difference-Abstractions (KVDA)} detects three non-trivial abstractions. In this example, circles represent states, arrows represent deterministic state-transitions and arrow annotations denote the immediate transition reward. All actions or states that are intersected by a red ellipse will be abstracted by KVDA.
}
  \label{fig:kvda_illustration}
\end{figure}

In this work we propose Known Value Differences Abstractions UCT (KVDA-UCT), that relaxes the strict abstraction conditions of OGA-UCT to detect almost as many abstractions as $(\varepsilon_{\text{a}},0)$-OGA in deterministic environments but without losing the exactness condition. The novel idea that makes this possible is to deliberately group states or state-action pairs that do not have the same value if we know the difference between their values which is inferred by analysis of the search tree's immediate rewards. When the abstractions are used, instead of averaging state-action pair values directly, their difference-accounted values are averaged.
The contributions of this paper can be summarized as follows:

\noindent \textbf{1.} We introduce the Known-Value-Difference (KVDA) abstraction framework that extends the Abstractions of State-Action Pairs (ASAP) framework used by OGA-UCT. Fig.~\ref{fig:kvda_illustration} is an example of a simple state-transition graph in which KVDA finds three non-trivial abstractions, while ASAP would detect none.

\noindent \textbf{2.} We propose and empirically evaluate KVDA-UCT, a modification of OGA-UCT that introduces no parameters and uses and builds KVDA abstractions. We show that KVDA-UCT outperforms OGA-UCT in most of the here-considered deterministic environments. We also compare KVDA-UCT with \mbox{$(\varepsilon_{\text{a}},0)$-OGA} and show that it either performs equally well or better than a parameter-optimized $(\varepsilon_{\text{a}},0)$-OGA agent. 

\noindent \textbf{3.} We also consider the stochastic setting where we generalize KVDA-UCT to $\varepsilon_{\text{t}}$-KVDA which allows for errors in the transition function when building the abstraction. Furthermore, we compare it to $(\varepsilon_{\text{t}},\varepsilon_{\text{a}})$-OGA and show that, unlike in the deterministic setting, KVDA rarely performs better than $(\varepsilon_{\text{t}},\varepsilon_{\text{a}})$-OGA in stochastic environments.

The paper is structured as follows. In \textbf{Section} \ref{sec:foundations}, the theoretical groundwork for automatic state and state-action pair abstractions is laid. In particular, we define OGA-UCT and $(\varepsilon_{\text{a}},\varepsilon_{\text{t}})$-OGA. Then, in \textbf{Section} \ref{sec:method}, our novel KVDA framework and both KVDA-UCT and $\varepsilon_{\text{t}}$-KVDA are introduced. Afterwards, in \textbf{Section} \ref{sec:experiment_setup}, the experimental setup is defined, then in \textbf{Section} \ref{sec:experiments}, the experimental results using this setup are shown and discussed. The paper is concluded by a discussion of the limitations of KVDA-UCT and avenues for future work in \textbf{Section} \ref{sec:future_work}.

%% file: sections/foundations.tex
In this section, we will be laying the theoretical groundwork for this paper as well as introducing related work. 
For a comprehensive overview of non-learning-based abstractions, we refer to the survey paper by Schmöcker and Dockhorn \citep{mysurvey}.

\noindent \textbf{Problem model and optimization objective:}
For our purposes, finite Markov Decision Processes \citep{sutton2018reinforcement} are used as the model for sequential, perfect-information decision-making tasks. $\Delta(X)$ denotes the probability simplex of a finite, non-empty set $X$ and the power set of $X$ is denoted by $\mathcal{P}(X)$.

\textit{Definition}:
    An \textit{MDP} is a 6-tuple $(S,\mu_0,T,\mathbb{A},\mathbb{P}, R)$ where the components are as follows:
    \begin{itemize}
        \item $S \neq \emptyset$ is the finite set of states.
        \item $\mu_0 \in \Delta(S)$ is the probability distribution for the initial state.
        \item $\emptyset \neq T \subsetneqq S$ is the (possibly empty) set of terminal states.
        \item $\mathbb{A}\colon (S \setminus T) \mapsto A$ maps each state $s$ to the available actions $\emptyset \neq \mathbb{A}(s) \subseteq A$ at state $s$ where $|A| < \infty$.
        \item $\mathbb{P}\colon (S \setminus T) \times A \mapsto \Delta(S )$ is the stochastic transition function where $\mathbb{P}(s^{\prime} |\: s,a)$ is used to denote the probability of transitioning from $s \in (S \setminus T)$ to $s^{\prime} \in S$ after taking action $a \in \mathbb{A}(s)$ in $s$.
        \item $R \colon (S \setminus T) \times A \mapsto \mathbb{R}$ is the reward function.
    \end{itemize}

\noindent Let $M = (S,\mu_0,T, \mathbb{A},\mathbb{P}, R)$ be an MDP. We define
    \mbox{$P \coloneqq \{(s,a)\: | \: s \in (S \setminus T), a \in \mathbb{A}(s)\}$}
as the set of all legal state-action pairs. The objective is to find a mapping (i.e. an agent) 
    $\pi \colon S \mapsto \Delta(A)$
such that $\pi$ maximizes the expected episode's return where the (discounted) return of an episode $s_0,a_0,r_0, \dots, s_n,a_n,r_n,s_{n+1}$ with $s_{n+1} \in T$ is given by $\gamma^0 r_0 + \ldots + \gamma^n r_n$.

\noindent \textbf{Abstractions of State-Action Pairs (ASAP):}
\label{sec:asap}
For MCTS-based abstraction research, the goal has been to detect state-action pairs with the same $Q^*$ value (the value under subsequent optimal play) in the search graph to increase sample efficiency by an intra-layer information flow \citep{uctJiang,AnandGMS15,OGAUCT}. In general, by abstractions of either the states of state-action-pairs we refer to equivalence relations over the state set $S$ or state-action pair set $P$. The equivalence classes are abstract states or state-action pairs.

The current state of the art is the Abstraction of State-Action Pairs in UCT (ASAP) abstraction framework \citep{AnandGMS15} that proposes rules to detect value-equivalent states and state-action pairs given an MDP transition graph and applies it to the current MCTS search graph (for details on MCTS, see Section \ref{sec:mcts}). The core idea of ASAP is to alternatingly construct a state abstraction given a state-action pair abstraction and a state-action pair abstraction given a state abstraction. 

Assume one is given a state abstraction $\mathcal{E}^{\prime} \subseteq S \times S$. The corresponding ASAP state-action pair abstraction $\mathcal{H} \subseteq P \times P$ is defined as grouping those state-action pairs with the same immediate reward and equal abstract successor distribution. Concretely,
any state-action-pair $(s_1,a_1),(s_2,a_2)$ is equivalent i.e. $((s_1,a_1),(s_2,a_2)) \in \mathcal{H}$ if and only if
\begin{equation}
    \begin{aligned}
         \quad F_{\text{a}} \coloneqq | R(s_1,a_1) -  R(s_2,a_2) |  
        & = 0 \\
        F_{\text{t}} \coloneqq \quad \sum \limits_{x \in \mathcal{X}} \bigg| \sum \limits_{s^{\prime} \in x} 
        \mathbb{P}(s^{\prime}|\: s_1,a_1) - \mathbb{P}(s^{\prime}|\: s_2,a_2) \bigg| 
       & = 0,
    \end{aligned}
    \label{eq:asap_distances}
\end{equation}
where $\mathcal{X}$ are the equivalence classes of $\mathcal{E}^{\prime}$. And given a state-action pair abstraction $\mathcal{H}^{\prime} \subseteq P \times P$, the corresponding ASAP state abstraction $\mathcal{E}$ groups all states whose actions can be mapped to each other, concretely: 
\begin{equation}
    \begin{aligned}
   & (s_1,s_2) \in \mathcal{E}  \iff \\
        &\forall a_1 \in \mathbb{A}(s_1) \, \exists a_2 \in \mathbb{A}(s_2):  
        ((s_1,a_1),(s_2,a_2)) \in \mathcal{H}_{i+1} \\
        &\forall a_2 \in \mathbb{A}(s_2) \, \exists a_1 \in \mathbb{A}(s_1):  
        ((s_1,a_1),(s_2,a_2)) \in \mathcal{H}_{i+1}.
    \end{aligned}
    \label{eq:asap_framework}
\end{equation}
To obtain the ASAP abstraction for a given MDP, these two constructing steps are repeated alternatingly until convergence.

\noindent \textbf{OGA-UCT:} \cite{OGAUCT} proposed OGA-UCT, which builds an ASAP-like abstraction in parallel to running MCTS. When building the abstraction OGA starts with the initial state abstraction that groups all terminal states of the same layer and puts the remaining states in their own singleton equivalence class. Furthermore, when building the ASAP abstraction on the current search graph, OGA ignores non-yet-sampled successors of state-action pairs that appear in Equation \ref{eq:asap_framework}. To make the frequent recomputation of the ASAP abstraction feasible, OGA keeps track of a recency counter for each Q-node and once it surpasses a certain threshold, recomputes its abstraction. If the abstraction changed, the parent states are recomputed too (and possibly their Q-node parents if their abstraction changed). By only locally checking for errors in the abstraction, OGA is able to keep track of an ASAP-like abstraction that is always close to the true ASAP abstraction of the current search tree.

The only MCTS component that OGA-UCT affects is the tree policy which is enhanced by using the aggregate returns and visits of a Q-node's abstract node to enhance the UCB value.
\\ \\
\noindent $\bm{(\varepsilon_{\text{a}},\varepsilon_{\text{t}})}$\textbf{-OGA}: The ASAP framework groups only value equivalent states and state-action pairs. This condition can be relaxed like already done by a predecessor of OGA-UCT, called AS-UCT \citep{uctJiang}, by allowing the rewards in Equation~\ref{eq:asap_distances} to differ by some threshold $\varepsilon_{\text{a}} > 0$ and the transition error $F_{\text{t}}$ of Equation~\ref{eq:asap_distances} to lie in the interval $[0,\varepsilon_{\text{t}}]$ where $0 \leq \varepsilon_{\text{t}} \leq 2$ is another parameter \textit{which does not have any effect in deterministic environments}. Since in general, positive threshold values do not induce an equivalence relation over state-action pairs, OGA-UCT has to be slightly modified to accommodate these approximate abstractions. This has been done by Schmöcker and Dockhorn \citep{ogacad} who introduced $(\varepsilon_{\text{a}},\varepsilon_{\text{t}})$-OGA.

\noindent \textbf{Auther automatic abstraction algorithm:}
The ASAP framework is the direct successor of Abstraction of States (AS) by \cite{uctJiang} that abstracts states if and only if their actions are pairwise similar in the sense that Equation \ref{eq:asap_framework} only has to be approximately satisfied as described in the $(\varepsilon_{\text{a}},\varepsilon_{\text{t}})$-OGA section above. While this paper focuses on deterministic domains, $(\varepsilon_\text{a},\varepsilon_{\text{t}})$-OGA (and related algorithms) have been further improved for stochastic settings by defining intra-abstraction policies \citep{intra}, or dynamically abandoning the abstraction \citep{EMCTSXu,ogacad}.

Yet another paradigm is demonstrated by \cite{HostetlerFD15} and \cite{aupo} who optimistically construct abstractions by starting with a very coarse abstraction (e.g. grouping everything together) and then they refine this abstraction, for example by repeatedly splitting large node groups in half \citep{HostetlerFD15}.

Research effort has also been dedicated towards automatic abstractions of the transition function, which on an abstract level can be described as pruning certain successors from the transition function \citep{SokotaHAK21,YoonFGK08,YoonFG07,saisubramanian2017optimizing}.

%% file: sections/method/motivation.tex
This section introduces our novel KVDA-UCT algorithm. First, we will provide an example of a concrete search graph in which ASAP misses abstractions that the Known Value Differences Abstractions (KVDA) framework would find which will be introduced after the search graph example. At the end of this section, we will be describing how KVDA is integrated into UCT to yield KVDA-UCT.

\subsection{Which abstractions ASAP misses}
Consider the state-transition graph that is illustrated in Fig.~\ref{fig:kvda_illustration} that consists of four states and four actions. The ASAP framework would not detect any equivalences. In fact, any framework that aims at finding value-equivalent states or state-action pairs would at most be able to detect that the two root actions are value-equivalent.
However, by analysing the state graph, one could derive that the $Q^*$ values of the action of node 2 differs only by $1$ from the $Q^*$ value of the action of node $3$. This holds true even if the state-graph would extend past node $4$. Consequently, the $V^*$ values of nodes 2 and 3 differ only by $1$. This in turn implies that the two actions of node 1 must have the same $Q^*$ value. This is the core idea behind our \textbf{Known-Value-Difference-Abstractions (KVDA)} framework which is to abstract states and state-action pairs if one can derive their value differences. Later, when using these abstractions to enhance MCTS, the differences only have to be subtracted when aggregating values. When viewed from the lenses of the KVDA framework, ASAP only groups states or state-action pairs with a value difference of $0$, i.e. detects only true equivalences.
Next, we will formalize the KVDA framework which both in theory and in our empirical evaluations (see Tab.~\ref{tab:abs_rates}, more details are given in the experimental section) detects strictly more abstractions.

%% file: sections/method/kvda_framework.tex
\subsection{The Known-Value-Difference-Abstractions framework}
Our method, the Known-Value-Difference-Abstractions (KVDA) extends the ASAP definition by additionally grouping states or state-action-pairs whose $V^*$ or $Q^*$ difference is known. While ASAP iteratively builds abstractions on abstractions, KVDA bootstraps of an abstraction, difference-function pair. More concretely, given a state abstraction $\mathcal{E}^{\prime}$ (or a state-action pair abstraction $\mathcal{H}^{\prime}$) and a difference function $d_{\text{s}}^{\prime} \colon S \times S \mapsto \mathbb{R}$ (or $d_{\text{a}}^{\prime} \colon P \times P \mapsto \mathbb{R}$), both a state-action-pair abstraction $\mathcal{H}$ (or a state abstraction $\mathcal{E}$) is produced as well as a state-action pair difference function $d_{\text{a}} \colon P \times P \mapsto \mathbb{R}$ (or a state difference function $d_{\text{s}}\colon S\times S\mapsto \mathbb{R}$). Both $d_{\text{a}}$ and $d_{\text{s}}$ will be constructed such that 
\begin{equation}
    \begin{aligned}
   d_{\text{a}}(p_1,p_2) & = d^*_{\text{a}}(p_1,p_2) \coloneqq Q^*(p_2) - Q^*(p_1) \\
       d_{\text{s}}(s_1,s_2) & = d^*_{\text{s}}(s_1,s_2) \coloneqq V^*(s_2) - V^*(s_1) \\
    \end{aligned}
    \label{eq:soundness}
\end{equation}
for any pair of states or state-action pairs in the same equivalence class (i.e., in the same abstract state). Even though in the experimental section, we will mostly consider deterministic environments, the now-to-be-described KVDA framework will be applicable to any MDP, including stochastic ones. 
Next, starting with the base case, we will formalize how KVDA abstractions are built.
\\ \\
\noindent \textbf{The base case:} Like ASAP, KVDA groups all terminal states into the same initial abstract node, the remaining states are singleton abstract nodes. The difference function between all nodes in the terminal abstract node is initialized with $0$.
\\ \\
\noindent \textbf{State-action-pair abstractions:} 
Using the same notation as in Section \ref{sec:foundations}, let $\mathcal{E}^{\prime} \subseteq S \times S$ be a state abstraction and \mbox{$d_{\text{s}}^{\prime}\colon S\times S \mapsto \mathbb{R}$} be a difference function. The corresponding KVDA state-action pair abstraction $\mathcal{H} \subseteq P \times P$ is defined as follows:
Any state-action pair $(s_1,a_1),(s_2,a_2)$ is equivalent i.e. $((s_1,a_1),(s_2,a_2)) \in \mathcal{H}$ if and only if
\begin{equation}
        \quad \sum \limits_{x \in \mathcal{X}} \bigg| \sum \limits_{s^{\prime} \in x} 
        \mathbb{P}(s^{\prime}|\: s_1,a_1) - \mathbb{P}(s^{\prime}|\: s_2,a_2) \bigg| 
        = 0,
\end{equation}
where $\mathcal{X}$ are the equivalence classes of $\mathcal{E}^{\prime}$. Note that this is almost identical to the ASAP definition except that the immediate rewards do not have to coincide. The difference function $d_{\text{a}}\colon P \times P \mapsto \mathbb{R}$ for two $(p_1,p_2) \in \mathcal{H}$ is given by
\begin{equation}
    \begin{aligned}
    d_{\text{a}}(p_1,p_2) = &  R(p_2) - R(p_1) \\
    + &   \sum \limits_{x \in \mathcal{X}} \sum \limits_{s^{\prime} \in x} 
        \left(\mathbb{P}(s^{\prime}|\: p_1) -   \mathbb{P}(s^{\prime}|\: p_2) \right) \cdot d_{\text{s}}^{\prime}(s^{\prime},s_{x}) 
    \end{aligned}
\end{equation}
where for each $x \in \mathcal{X}$, $s_{x}$ is an arbitrarily chosen but fixed representative of $\mathcal{X}$. We will later see that the value of $d_{\text{a}}$ is independent of this choice.
\\ \\
\noindent \textbf{State abstractions:}
Given a state-action pair abstraction $\mathcal{H}^{\prime} \subseteq P \times P$ and a state-action pair difference function $d_{\text{a}}\colon P \times P \mapsto \mathbb{R}$ the corresponding KVDA state abstraction $\mathcal{E} \subseteq S \times S$ groups all states whose actions can be mapped to each other, and whose mappings all have the same value difference: 
\begin{equation}
    \begin{aligned}
   & (s_1,s_2) \in \mathcal{E}  \iff \exists d \in \mathbb{R}: \\
        &\forall a_1 \in \mathbb{A}(s_1) \, \exists a_2 \in \mathbb{A}(s_2):  
        ((s_1,a_1),(s_2,a_2)) \in \mathcal{H}^{\prime}\\
        & \land d_{\text{a}}((s_1,a_1),(s_2,a_2)) = d \\
        &\forall a_2 \in \mathbb{A}(s_2) \, \exists a_1 \in \mathbb{A}(s_1):  
        ((s_1,a_1),(s_2,a_2)) \in \mathcal{H}^{\prime} \\
        &\land d_{\text{a}}((s_2,a_2),(s_1,a_1)) = d.
    \end{aligned}
    \label{eq:kvda_states}
\end{equation}
The difference function $d_{\text{s}}(s_1,s_2)$ is defined as the value $d$ in the equation above.
\\ \\
\textbf{Theoretical guarantees: }

\textit{Convergence:}
Given an MDP, the above-described construction steps can be repeated until convergence, which is guaranteed as in our MDP definition, there are finitely many states and state-action pairs and each construction step either leaves the abstraction unchanged or reduces the number of equivalence classes. The abstraction that one obtains at convergence is called the KVDA abstraction of an MDP.

\textit{Soundness of $d_{\text{a}}$ and $d_{\text{s}}$: } Both $d_{\text{a}}$ and $d_{\text{s}}$ at convergence are equal to the differences of their arguments $Q^*$ or $V^*$ values as formulated in Eq. \ref{eq:soundness}. This is proven in the supplementary materials in Section \ref{sec:proof}.
\\ \\

%% file: sections/method/implementation.tex
\subsection[{KVDA-UCT and $\varepsilon_{\text{t}}$-KVDA}]{KVDA-UCT and $\bm{\varepsilon}_{\text{t}}$-KVDA}
In this section, we will describe how the KVDA abstraction framework is integrated into MCTS, which is fully analogous to how $(\varepsilon_{\text{a}},\varepsilon_{\text{t}})$-OGA (which itself is equivalent to OGA-UCT for \mbox{$\varepsilon_{\text{a}} = \varepsilon_{\text{t}} = 0 $}) integrates the ASAP framework. The usage of the following to-be-described modifications to $(\varepsilon_{\text{a}},\varepsilon_{\text{t}})$-OGA is called $\boldsymbol{\varepsilon_{\text{t}}}$\textbf{-KVDA}. For the case $\varepsilon_{\text{t}}=0$, the algorithm is simply called \textbf{KVDA-UCT}. $\boldsymbol{\varepsilon_{\text{t}}}$\textbf{-KVDA} does not depend on $\varepsilon_{\text{a}}$ unlike $(\varepsilon_{\text{a}},\varepsilon_{\text{t}})$-OGA.

\noindent\textbf{1)} Instead of (approximate) ASAP abstractions, (approximate) KVDA abstractions of the search tree are built. Both $(\varepsilon_{\text{a}},\varepsilon_{\text{t}})$-OGA and $\varepsilon_{\text{t}}$-KVDA allow the transition error $F_{\text{t}}$ of Equation $\ref{eq:asap_distances}$ to be in the interval $[0,\varepsilon_{\text{t}}]$.

\noindent\textbf{2)} In addition to abstract Q nodes, abstract state nodes now also keep track of a representative. In our case, this is the first original node added to the abstract node, and if that one is removed, a random new representative is chosen.

\noindent\textbf{3)} Each abstract Q node's value is tracked as the value of its representative $\mathcal{Q}_{\text{R}}$ that encodes the state-action-pair $p_{\text{R}} \in P$. When another original node $\mathcal{Q}$ of the same abstract node representing $p \in P$ is backed up in the MCTS backup phase with value $v \in \mathbb{R}$, the value $v + d_{\text{a}}(p,p_{\text{R}})$ is added to the abstract node's statistics. In turn, $\mathcal{Q}$ extracts its aggregated returns from the abstract node by subtracting $ d_{\text{a}}(p,p_{\text{R}})$ from the abstract node's value. If an abstract node's representative changes to $\mathcal{Q}^{\prime}_{\text{R}}$ encoding $p_{\text{R}}^{\prime} \in P$, then $n \cdot d_{\text{a}}(p_{\text{R}},p_{\text{R}}^{\prime})$ is added to the statistics where $n$ is the abstract visit count.

\noindent \textbf{4)} To reduce the computational load of finding a perfect match as the one required for the state abstractions in Equation~\ref{eq:kvda_states}, we check for a stricter condition for states $s_1,s_2$. It is first checked that within $s_1$ (and analogously $s_2$) all actions within the same abstract node have a value difference of zero. Then, it is tested for all abstract Q nodes of $s_1$ if the value difference between an arbitrarily chosen ground action of $s_1$ and one of $s_2$ is constant for all abstract nodes.  

\noindent \textbf{5)} Finally, whenever a Q-node's recency counter reaches the threshold, the value difference to its representative is recalculated. A change in this difference also results in a reevaluation (and subsequent recency counter reset) of the parent nodes' abstractions and difference functions.

%% file: sections/experiment_setup.tex
In this section, we describe the general experiment setup. Any deviations from this setup will be explicitly mentioned.

\textbf{Problem models:}
For this paper, we ran our experiments on a variety of MDPs, all of which are either from the International Probabilistic Planning Conference \citep{grzes2014ippc}, are well-known board games, or are commonly used in the abstraction algorithm literature. Since we will compare KVDA-UCT to OGA-UCT we chose only environments with a non-constant and non-sparse reward function, as constant reward environments would imply that KVDA-UCT and OGA-UCT are semantically equivalent.

All MDPs originally feature stochasticity, however, for some experiments we considered their deterministic versions which are obtained by sampling a single successor of each state-action pair. For each episode, new successors are sampled.
All experiments were run on the finite horizon versions of the considered MDPs with a default horizon of 50 steps and 200 for the board games with a planning horizon of 50 and a discount factor $\gamma=1$. 

The board games are transformed into MDPs by inserting a deterministic One-Step-Lookahead agent as the opponent. Furthermore, since they are all sparse-reward (i.e. either win or lose) they were also transformed into dense-reward MDPs by using heuristics. For each board game we defined a heuristic $V^{\text{h}}$ that assigns each state a heuristic value for its state value. The reward of the (deterministic) transition $s,a,s^{\prime}$ is then given by $V^{\text{h}}(s^{\prime}) - V^{\text{h}}(s)$. The concrete heuristics used along with a brief description of all MDPs is provided in the supplementary materials in Section~\ref{sec:problem_descriptions}. The deterministic MDPs are denoted by adding the prefix \textit{d-} and the stochastic ones are denoted by the prefix \textit{s-}.

\textbf{Parameters:}
Since the problem domains have vastly different reward scales, we use the dynamic exploration factor Global Std \citep{demcts} which has the form $C \cdot \sigma$ where $\sigma$ is the standard deviation of the Q values of all nodes in the search tree and $C \in \mathbb{R}^+$ is some fixed parameter. Furthermore, we always use $K=3$ as the recency counter which was proposed by \cite{OGAUCT}.

\textbf{Evaluation:}
Each experiment is repeated at least 2000 times and all confidence that we denote in the following are $99\%$ confidence intervals with range $\approx 2 \cdot 2.33$ times the standard error.

\textbf{Normalized pairings score}: We will later construct the \textit{normalized pairings score} to test the generalization capabilities of KVDA-UCT. This is the same score as used in \cite{demcts}. The pairings score is constructed as follows. Let $\{\pi_1,\dots,\pi_n\}$ be $n$ agents (e.g. each KVDA-UCT agent along with its parameter setting is an agent) where each agent was evaluated on $m$ tasks (later, a task will be a given MCTS iteration budget and an environment). This induces a matrix of size $n \times n$ with the entry $(i,j)$ being equal to the number of tasks where $\pi_i$ achieved a higher performance than $\pi_j$ subtracted by the number of times it performed worse, divided by $m$.
The normalized pairings score $s_i \leq i \leq n$ is then obtained by taking the average of $i$-th row when excluding the $i$-th column. 

\textbf{Reproducibility:}
For reproducibility, we released our implementation \citep{repo}. Our code was compiled with g++ version 13.1.0 using the -O3 flag (i.e. aggressive optimization). 

%% file: sections/experiments/experiments.tex
This section presents the experimental results of KVDA-UCT. We considered two settings. Firstly, in Section \ref{sec:abs_stats} we measured the number of additional abstractions KVDA-UCT finds in comparison to OGA-UCT. Then, we evaluated KVDA-UCT on deterministic settings in which the losslessness of the abstraction is guaranteed in Section~\ref{sec:param_optimized} and Section~\ref{sec:gen}. Lastly, we present the results for stochastic environments in Section~\ref{sec:stochastic}.

\noindent \textbf{Abstractions that KVDA-UCT finds but OGA-UCT does not:}
\label{sec:abs_stats}
Firstly, we empirically measured the number of non-trivial abstractions (i.e. those that aren't of size one) that KVDA-UCT, OGA-UCT, and $(\infty,0)$-UCT find. For all deterministic environments, Tab.~\ref{tab:abs_rates} denotes the average ratio of non-trivial abstract state-action pairs (synonymously abstract Q nodes) to the number of total abstract Q nodes in their respective search trees. In most environments, KVDA-UCT detects more abstractions than OGA-UCT, including environments where the abstraction rate more than doubles, such as SysAdmin or Wildfire. Furthermore, KVDA-UCT detects roughly as many abstractions as $(\infty,0)$-OGA in most environments, which fully ignores rewards when building abstractions.

\begin{table}[ht]
\centering
\caption{The average ratio of abstract Q nodes in KVDA, OGA-UCT's and $(\infty,0)$-OGA's respective search trees after 1000 iterations with $\lambda=2$ that encompass more than one original node divided by the total number of abstract Q nodes. The states in which these search trees are built are sampled from an OGA-UCT agent with $\lambda=2$ and 500 iterations. This measurement excludes all size-one abstract Q nodes whose original Q node has not yet reached the recency counter (see Section \ref{sec:asap}) required for the first abstraction update. Hence, a ratio of $1$ means that no abstractions were found while a ratio of $0$ means that every state-action pair has been abstracted with another. Note that KVDA-UCT (our method) finds more abstractions than OGA-UCT in nearly every environment and in general as many as $(\infty,0)$-OGA. While there are some environments where there is no gain, such as Constrictor, there are other environments such as Wildfire or SysAdmin, where the abstraction rate more than doubles.}
\label{tab:abs_rates}
\scalebox{0.75}{
\begin{tabular}{l c c c}
\hline
\textbf{Domain} & \shortstack{KVDA-UCT \\ (ours)} & \shortstack{OGA- \\ UCT} & \shortstack{$(\infty,0)$- \\ OGA} \\
\hline
d-Academic Advising & 0.69 & 0.73 & 0.68 \\
d-Connect4 & 0.80 & 0.91 & 0.80 \\
d-Constrictor & 0.98 & 0.97 & 0.97 \\
d-Cooperative Recon & 0.44 & 0.56 & 0.48 \\
d-Earth Observation & 0.65 & 0.99 & 0.68 \\
d-Elevators & 0.28 & 0.32 & 0.29 \\
d-Game of Life & 0.54 & 0.56 & 0.55 \\
d-Manufacturer & 0.64 & 0.95 & 0.79 \\
d-Othello & 0.96 & 0.99 & 0.96 \\
d-Pusher & 0.96 & 0.99 & 0.96 \\
d-Push Your Luck & 0.23 & 0.48 & 0.16 \\
d-Red Finned Blue Eye & 0.72 & 0.84 & 0.70 \\
d-Sailing Wind & 0.70 & 0.92 & 0.74 \\
d-Saving & 0.96 & 0.96 & 0.96 \\
d-Skills Teaching & 0.59 & 0.65 & 0.59 \\
d-SysAdmin  & 0.15 & 0.48 & 0.20 \\
d-Tamarisk & 0.35 & 0.56 & 0.39 \\
d-Traffic & 0.53 & 0.54 & 0.53 \\
d-Wildfire & 0.19 & 0.37 & 0.30 \\
d-Wildlife Preserve & 0.06 & 0.08 & 0.08 \\
\hline
\end{tabular}}
\end{table}

\begin{figure}[H]
\centering

\begin{minipage}{0.32\textwidth}
\centering
\includegraphics[width=\linewidth]{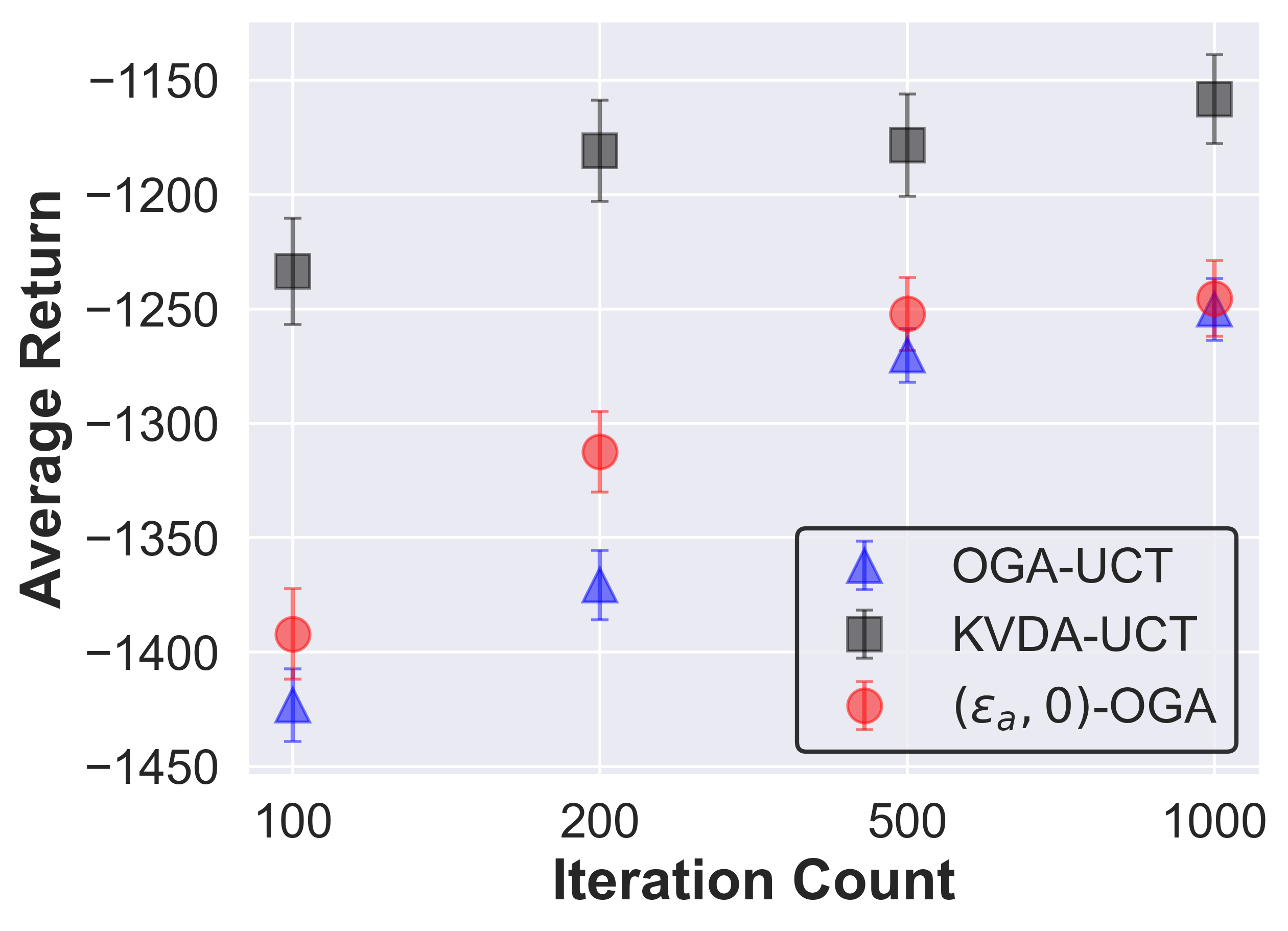}
\caption*{(a) d-Manufacturer}
\end{minipage}
\begin{minipage}{0.32\textwidth}
\centering
\includegraphics[width=\linewidth]{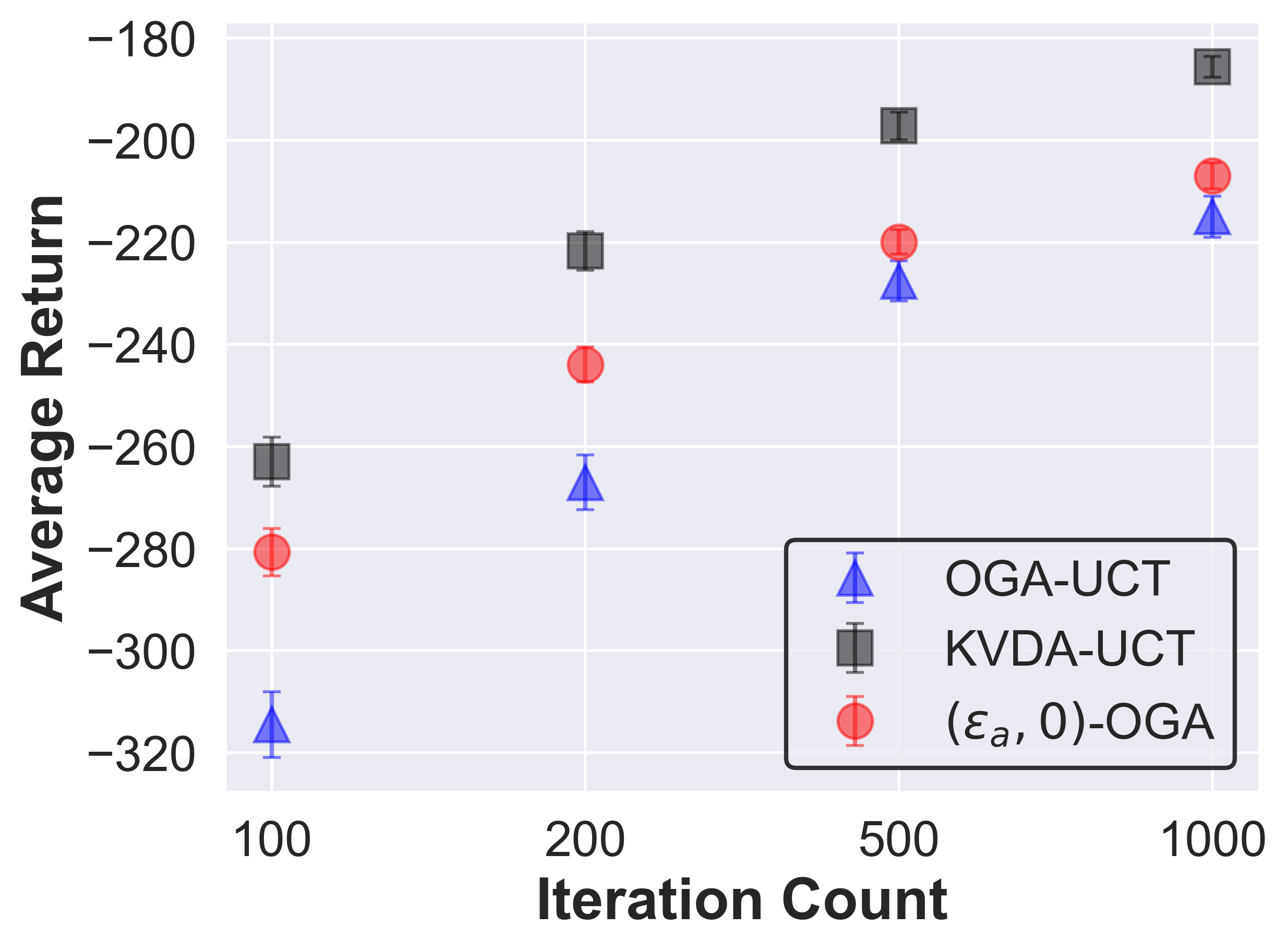}
\caption*{(b) d-Tamarisk}
\end{minipage}
\begin{minipage}{0.32\textwidth}
\centering
\includegraphics[width=\linewidth]{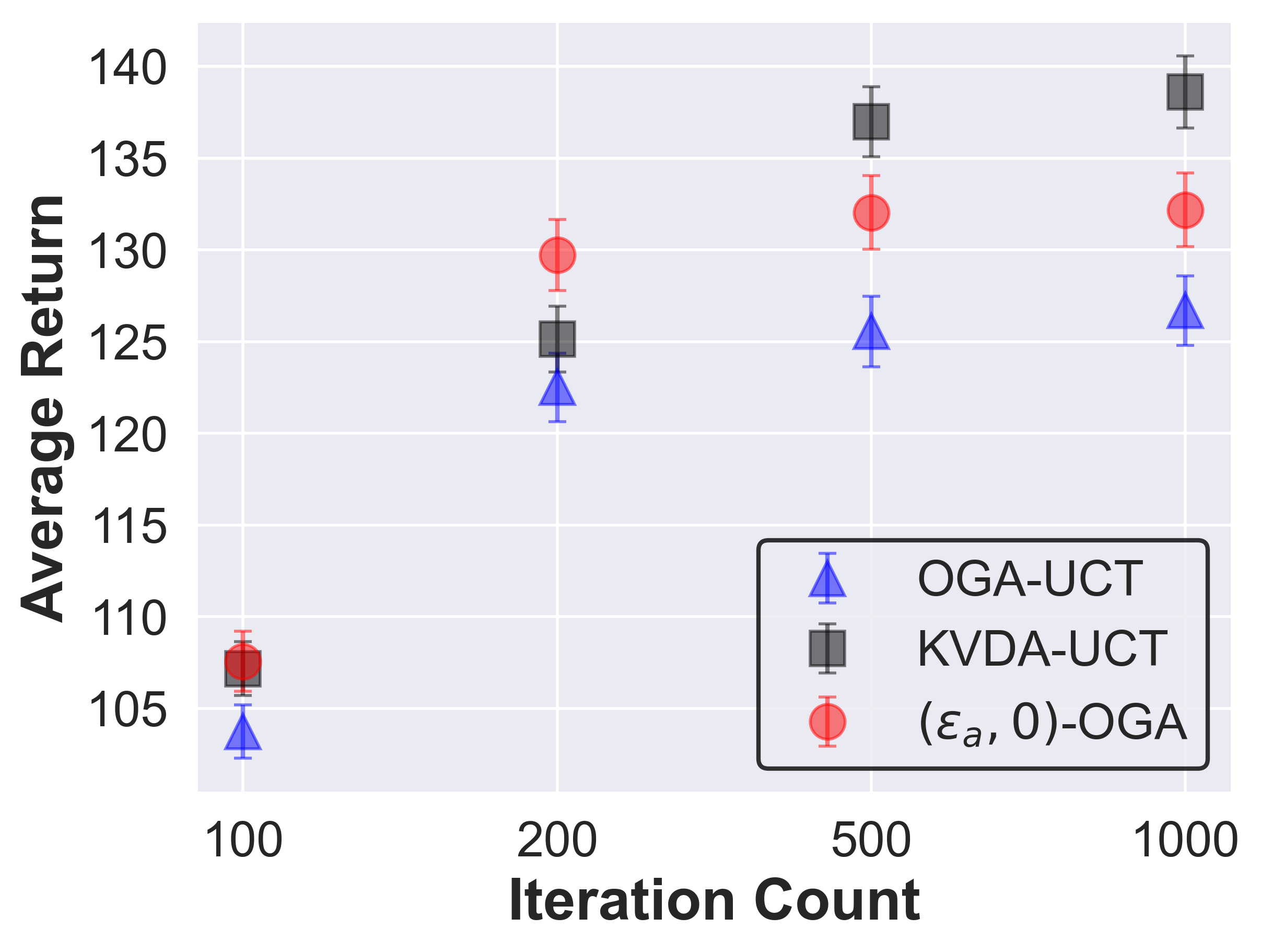}
\caption*{(c) d-Push Your Luck}
\end{minipage}

\caption{The performance plots in dependence of the iteration budget of parameter-optimized KVDA-UCT (our method), OGA-UCT \citep{OGAUCT}, and $(\varepsilon_{\text{a}},0)$-OGA on selected environments. KVDA-UCT clearly outperforms its competitors. The plots for the remaining environments can be found in the supplementary materials in Section~\ref{sec:performance_graphs}.}
\label{fig:performances}
\end{figure}

\noindent \textbf{Parameter-optimized KVDA-UCT:}
\label{sec:param_optimized}
Next, we compared KVDA-UCT, $(0,0)$-OGA (i.e. standard OGA-UCT), and $(\varepsilon_{\text{a}},0)$-OGA with \mbox{$\varepsilon_{\text{a}} > 0$} in terms of their parameter-optimized performances on deterministic environments.
For all methods, we optimized over \mbox{$C \in \{0.5,1,2,4,8,16\}$} and some domain-specific values for $\varepsilon_{\text{a}}$ that are listed in the supplementary materials in Tab.~\ref{tab:epsa_values}. Each parameter combination was evaluated on all here-considered deterministic environments with a budget of 100, 200, 500, and 1000 MCTS iterations. 
Tab.~\ref{tab:combined_param_optimized} list the performance for 100 and 1000 iterations, the tables for 200 and 500 iterations are found in the supplementary materials in Tab.~\ref{tab:param_optimized_others200} and ~\ref{tab:param_optimized_others500}. The results clearly show that KVDA either outperforms all other competitor methods or is tightly within the confidence bounds. There are a number of environments such as Manufacturer, Tamarisk, or Push Your Luck where KVDA-UCT simultaneously outperforms all competitor methods at once. Figure \ref{fig:performances} plots the performances in dependence of the iteration budget for these tasks. The performance graphs for all other environments are found in the supplementary materials in Section \ref{sec:performance_graphs}.

\begin{table*}
\centering
\caption{Average returns $(\uparrow)$ using 100 (right) and 1000 (left) MCTS iterations for KVDA-UCT, $(\infty,0)$-OGA, $(0,0)$-OGA (i.e. OGA-UCT), and the maximal performance of $(\varepsilon_{\text{a}},0)$-OGA when varying $\varepsilon_{\text{a}} > 0$. In the supplementary materials in Tab.\ref{tab:epsa_values} the domain-specific $\varepsilon_{\text{a}}$-values are listed. All performances are the maximal performances obtained by varying the exploration constants $C \in \{0.5,1,2,4,8,16\}$. Note that KVDA-UCT (our method) outperforms OGA-UCT in most environments. Furthermore, KVDA-UCT is either equal or performs better than parameter-optimized $(\varepsilon_{\text{a}},0)$-OGA even though KVDA introduces no extra parameter.}
\label{tab:combined_param_optimized}
\scalebox{0.58}{
\setlength{\tabcolsep}{1mm}
\begin{tabular}{l|c c c c|c c c c}
\toprule

& \multicolumn{4}{c|}{\textbf{1000 Iterations}} 
& \multicolumn{4}{c}{\textbf{100 Iterations}} \\
\cmidrule(lr){2-5} \cmidrule(lr){6-9}
\textbf{Domain}  & \shortstack{$(0,0)$-\\OGA} 
& \shortstack{$(\infty,0)$-\\OGA} 
& \shortstack{$(\varepsilon_{\text{a}},0)$-\\OGA} 
& \shortstack{\textbf{KVDA}\\(ours)} 
& \shortstack{$(0,0)$-\\OGA} 
& \shortstack{$(\infty,0)$-\\OGA} 
& \shortstack{$(\varepsilon_{\text{a}},0)$-\\OGA} 
& \shortstack{\textbf{KVDA}\\(ours)} \\
\midrule
d-Connect4 & $42.7 \pm 0.6$ & $46.8 \pm 1.0$ & $47.5 \pm 0.9$ & $\boldsymbol{47.9 \pm 0.6}$ & $28.3 \pm 0.3$ & $\boldsymbol{28.8 \pm 0.5}$ & $\boldsymbol{28.8 \pm 0.5}$ & $28.4 \pm 0.4$ \\
d-Constrictor & $96.1 \pm 0.3$ & $95.0 \pm 0.2$ & $\boldsymbol{96.1 \pm 0.3}$ & $96.0 \pm 0.3$ & $85.0 \pm 0.6$ & $84.1 \pm 0.4$ & $84.9 \pm 0.9$ & $\boldsymbol{85.1 \pm 0.6}$ \\
d-Othello & $\boldsymbol{181.2 \pm 1.2}$ & $160.2 \pm 0.8$ & $180.7 \pm 1.5$ & $179.6 \pm 1.2$ & $154.7 \pm 1.3$ & $146.8 \pm 0.8$ & $\boldsymbol{155.1 \pm 1.8}$ & $154.8 \pm 1.3$ \\
d-Pusher & $\boldsymbol{104.5 \pm 0.0}$ & $\boldsymbol{104.5 \pm 0.0}$ & $\boldsymbol{104.5 \pm 0.0}$ & $\boldsymbol{104.5 \pm 0.0}$ & $\boldsymbol{104.5 \pm 0.0}$ & $\boldsymbol{104.5 \pm 0.0}$ & $\boldsymbol{104.5 \pm 0.0}$ & $\boldsymbol{104.5 \pm 0.0}$ \\
d-Academic Advising & $\boldsymbol{-39.9 \pm 0.2}$ & $-40.0 \pm 0.2$ & $-40.0 \pm 0.2$ & $-40.1 \pm 0.2$ & $-56.6 \pm 0.7$ & $-56.2 \pm 0.6$ & $-56.2 \pm 0.6$ & $\boldsymbol{-56.1 \pm 0.6}$ \\
d-Cooperative Recon & $16.1 \pm 0.1$ & $14.8 \pm 0.1$ & $\boldsymbol{16.2 \pm 0.1}$ & $16.0 \pm 0.1$ & $10.8 \pm 0.3$ & $10.9 \pm 0.3$ & $11.0 \pm 0.3$ & $\boldsymbol{11.0 \pm 0.3}$ \\
d-Earth Observation & $-7.18 \pm 0.11$ & $-30.0 \pm 0.4$ & $-30.0 \pm 0.4$ & $\boldsymbol{-7.02 \pm 0.10}$ & $-10.2 \pm 0.2$ & $-28.4 \pm 0.2$ & $-28.4 \pm 0.2$ & $\boldsymbol{-7.45 \pm 0.12}$ \\
d-Elevators & $-14.9 \pm 0.4$ & $-14.2 \pm 0.3$ & $-14.0 \pm 0.3$ & $\boldsymbol{-13.9 \pm 0.4}$ & $-18.0 \pm 0.3$ & $-18.1 \pm 0.3$ & $\boldsymbol{-17.9 \pm 0.3}$ & $-18.0 \pm 0.3$ \\
d-Game of Life & $\boldsymbol{666.7 \pm 2.5}$ & $664.6 \pm 2.5$ & $665.8 \pm 2.5$ & $664.8 \pm 2.6$ & $641.8 \pm 2.1$ & $\boldsymbol{642.1 \pm 2.2}$ & $\boldsymbol{642.1 \pm 2.2}$ & $641.9 \pm 2.1$ \\
d-Manufacturer & $-1255.6 \pm 15.0$ & $-1658.4 \pm 22.1$ & $-1246.0 \pm 16.2$ & $\boldsymbol{-1158.2 \pm 19.4}$ & $-1423.1 \pm 15.9$ & $-1680.3 \pm 22.4$ & $-1392.6 \pm 20.2$ & $\boldsymbol{-1233.5 \pm 23.2}$ \\
d-Push Your Luck & $125.1 \pm 1.9$ & $66.7 \pm 1.1$ & $132.4 \pm 2.5$ & $\boldsymbol{137.9 \pm 2.2}$ & $103.5 \pm 1.5$ & $66.3 \pm 0.9$ & $107.0 \pm 1.6$ & $\boldsymbol{107.6 \pm 1.5}$ \\
d-RedFinnedBlueEye & $8191.3 \pm 44.7$ & $7930.0 \pm 44.7$ & $7950.6 \pm 33.5$ & $\boldsymbol{8229.8 \pm 46.5}$ & $7683.4 \pm 33.8$ & $7464.4 \pm 34.7$ & $7495.1 \pm 35.1$ & $\boldsymbol{7698.2 \pm 34.0}$ \\
d-Sailing Wind & $-40.0 \pm 0.6$ & $-39.1 \pm 0.6$ & $-38.9 \pm 0.6$ & $\boldsymbol{-37.7 \pm 0.6}$ & $-64.7 \pm 0.8$ & $-64.9 \pm 0.9$ & $-64.9 \pm 0.9$ & $\boldsymbol{-63.6 \pm 0.8}$ \\
d-Saving & $\boldsymbol{66.0 \pm 0.2}$ & $63.0 \pm 0.3$ & $65.4 \pm 0.2$ & $65.4 \pm 0.2$ & $57.1 \pm 0.2$ & $55.2 \pm 0.3$ & $56.8 \pm 0.2$ & $\boldsymbol{57.2 \pm 0.2}$ \\
d-Skills Teaching & $207.9 \pm 4.6$ & $211.3 \pm 5.1$ & $211.3 \pm 5.1$ & $\boldsymbol{216.2 \pm 4.5}$ & $159.0 \pm 3.9$ & $158.3 \pm 3.9$ & $160.7 \pm 3.8$ & $\boldsymbol{162.3 \pm 3.8}$ \\
d-SysAdmin & $477.1 \pm 1.5$ & $448.4 \pm 1.3$ & $450.7 \pm 1.1$ & $\boldsymbol{477.2 \pm 1.5}$ & $475.5 \pm 1.7$ & $449.5 \pm 1.2$ & $450.1 \pm 1.2$ & $\boldsymbol{479.1 \pm 1.4}$ \\
d-Tamarisk & $-214.5 \pm 3.9$ & $-208.2 \pm 2.3$ & $-206.4 \pm 2.7$ & $\boldsymbol{-185.6 \pm 2.0}$ & $-315.8 \pm 6.5$ & $-285.4 \pm 4.8$ & $-284.9 \pm 4.7$ & $\boldsymbol{-263.0 \pm 4.8}$ \\
d-Traffic & $\boldsymbol{-1.54 \pm 0.07}$ & $-1.61 \pm 0.07$ & $-1.61 \pm 0.07$ & $-1.58 \pm 0.07$ & $\boldsymbol{-9.21 \pm 0.22}$ & $\boldsymbol{-9.21 \pm 0.22}$ & $\boldsymbol{-9.21 \pm 0.22}$ & $\boldsymbol{-9.21 \pm 0.22}$ \\
d-Wildfire & $-195.6 \pm 55.9$ & $-503.5 \pm 36.1$ & $-415.0 \pm 36.1$ & $\boldsymbol{-179.9 \pm 49.2}$ & $-194.1 \pm 36.1$ & $-498.7 \pm 36.1$ & $-408.9 \pm 36.1$ & $\boldsymbol{-173.2 \pm 36.1}$ \\
d-Wildlife Preserve & $1388.0 \pm 0.8$ & $1387.9 \pm 0.7$ & $1387.9 \pm 0.7$ & $\boldsymbol{1388.5 \pm 0.7}$ & $\boldsymbol{1388.9 \pm 0.9}$ & $\boldsymbol{1388.9 \pm 0.9}$ & $\boldsymbol{1388.9 \pm 0.9}$ & $\boldsymbol{1388.9 \pm 0.9}$ \\
\bottomrule
\end{tabular}
}
\end{table*}

\begin{figure*}[h!]
\centering
\includegraphics[width=1\textwidth]{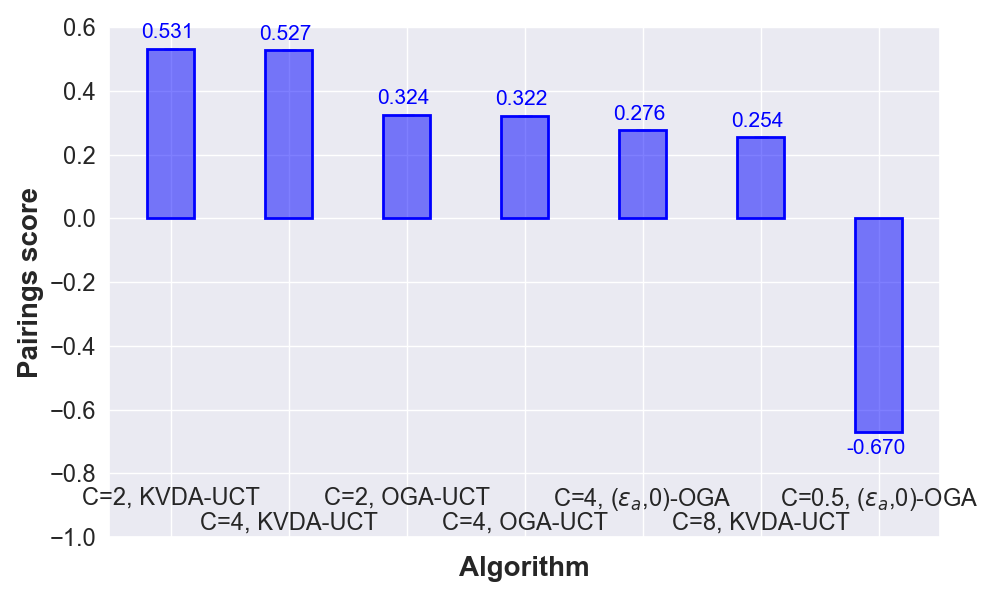}
\caption{The normalized pairings score $(\uparrow)$ for the top 6 and the worst agent on deterministic environments. The agents considered were KVDA-UCT (our method) which performs best overall, OGA-UCT \citep{OGAUCT}, and $(\varepsilon_{\text{a}},0)$-OGA \citep{ogacad}, $\varepsilon_{\text{a}} > 0$ with the exploration constants $C \in \{0.5,1,2,4,8,16\}$ and budgets of $\{100,200,500,1000\}$ iterations. The top two spots are occupied by our method KVDA-UCT, with the best overall performing algorithm being KVDA-UCT with $C=2$.}
\label{fig:score_gen}
\end{figure*}

\noindent \textbf{Single-parameter KVDA-UCT:}
\label{sec:gen}
Next, the generalization capabilities of KVDA-UCT in comparison to OGA-UCT and $(\varepsilon_{\text{a}},0)$-OGA on deterministic environments were tested. Like last section, we ran all algorithms using 100, 200, 500, and 1000 MCTS iterations whilst varying $C \in \{0.5,1,2,4,8,16\}$ and $\varepsilon_{\text{a}}$ with domain-specific values (see supplementary materials Tab.~\ref{tab:epsa_values}). Instead of considering the best-performances, we calculated the normalized pairings score for each parameter combination (except for $(\varepsilon_{\text{a}},0)$-OGA where we considered the maximum performance across all $\varepsilon_{\text{a}} > 0 $ values a single parameter combination), constructed over the tasks which are the pairs of all environments and iteration budgets. Bar chart~\ref{fig:score_gen} shows the 6 agents with the highest score as well as the agent with the lowest score. Both the top spots are occupied by our KVDA-method, followed by OGA-UCT and $(\varepsilon_{\text{a}},0)$-OGA. The best performing algorithm overall is KVDA-UCT with $C=4$.

\noindent \bm{$\varepsilon_{\text{t}}$}\textbf{-KVDA on stochastic environments:}
\label{sec:stochastic}
Lastly, the performance of $\varepsilon_{\text{t}}$-KVDA on stochastic environments were tested by running both $\varepsilon_{\text{t}}$-KVDA and $(\varepsilon_{\text{a}},\varepsilon_{\text{t}})$-OGA on the stochastic versions of the here-considered environments. We considered the iteration budgets of 100 and 1000 iterations and varied $C \in \{0.5,1,2,4,8,16\}$, $\varepsilon_{\text{t}} \in \{0,0.4,0.8,1.2,1.6\}$ and the same domain-specific $\varepsilon_{\text{a}}$ that were used in the previous sections whose values are listed in the supplementary materials in Tab.~\ref{tab:epsa_values}. For each environment, Tab.~\ref{tab:stochastic} of the supplementary materials lists the parameter-optimized performances of both algorithms. In contrast to the deterministic setting, KVDA does not outperform OGA. For most environments both perform equally well with some exceptions such as Manufacturer were $\varepsilon_{\text{t}}$-KVDA clearly performs best and Tamarisk were $(\varepsilon_{\text{a}},\varepsilon_{\text{t}})$-OGA performs best. We believe mediocre performance of $\varepsilon_{\text{t}}$-KVDA stems from the fact that since $\varepsilon_{\text{t}}$-KVDA essentially ignores immediate rewards when it builds abstractions, the number of faulty abstractions that occur in this approximate setting is increased. In the deterministic setting, no faulty abstractions are built.

%% file: sections/sum_concl_fw.tex
This paper introduced KVDA-UCT, an extension to OGA-UCT, that additionally groups states and state-action pairs that do not have the same $Q^*$ or $V^*$ value as long as the difference is known. We evaluated and compared KVDA-UCT to OGA-UCT and $(\varepsilon_{\text{a}},0)$-OGA on a deterministic setting where KVDA-UCT outperforms all its competitors both in terms of generalization as well as the parameter-optimized performance. This does not hold true for the stochastic setting for which we generalized KVDA-UCT to 
$\varepsilon_{\text{t}}$-KVDA which is only clearly better than $(\varepsilon_{\text{a}},\varepsilon_{\text{t}})$-OGA in few environments.

A first avenue for future work will be to further investigate the reasons for the mediocre performance of the experiments in this paper of the stochastic setting and develop an extension of $\varepsilon_{\text{t}}$-KVDA suited for this setting.
Another limitation of KVDA abstractions in its current form is its limitation to MDPs (i.e. single-player games). Even though MCTS is in principle applicable to multi-player games, KVDA is not because in any state, there is in general, no unique $V^*$ value from which differences can be built, as there are potentially many equilibria with different payoff vectors. Future work will be to extend KVDA to this setting.

%% file: sections/appendix.tex
\subsection[Proof that $d^*_{\text{a}} = d_{a}$ and $d^*_{s} = d_{s}$]{Proof that $d^*_{a} = d_{a}$ and $d^*_{s} = d_{s}$}
\label{sec:proof}
In this section, it will be shown that the KVDA difference functions $d_{\text{a}}$ and $d_{\text{s}}$ at any step (including the step of convergence)
coincide with $d^*_{\text{a}}$ and $d^*_{\text{s}}$ for state or state-action pairs within the same abstract state. This will be proven inductively. First, note that this statement is true for the base case as all terminal states have a value of $V^* = 0$; hence, their differences are also 0. For the following induction steps, assume that $d_{\text{a}}$ is bootstrapped by $d^{\prime}_{\text{s}}$ and $d_{\text{s}}$ is bootstrapped by $d^{\prime}_{\text{a}}$.
Next, let $(p_1,p_2) \in \mathcal{H}$. By definition, it holds that
\begin{equation}
    d^*(p_1,p_2) = R(p_2) - R(p_1) + \underbrace{\sum\limits_{s \in S} (\mathbb{P}(s |\: p_2 ) - \mathbb{P}(s|\:p_1)) V^*(s)}_{L \coloneqq}.
\end{equation}
By rewriting $V^*(s)$ in terms of the difference to its abstract representative, one obtains
\begin{equation}
    \begin{aligned}
         L  
        & = \sum \limits_{x \in \mathcal{X}}\sum \limits_{s^{\prime} \in x} 
        \left(\mathbb{P}(s^{\prime}|\: p_2) -   \mathbb{P}(s^{\prime}|\: p_1) \right) (V^*(s_x) - \underbrace{d^*_{\text{s}}(s^{\prime},s_x))}_{= d^{\prime}_{\text{s}} \text{ by induction}} \\
        & =  \sum \limits_{x \in \mathcal{X}}\sum \limits_{s^{\prime} \in x} 
        (\mathbb{P}(s^{\prime}|\: p_2) -   \mathbb{P}(s^{\prime}|\: p_1)) V^*(s_x) \\
        & + \sum \limits_{x \in \mathcal{X}}\sum \limits_{s^{\prime} \in x} 
        (\mathbb{P}(s^{\prime}|\: p_1) -   \mathbb{P}(s^{\prime}|\: p_2)) \cdot d^*_{\text{s}}(s^{\prime},s_x) \\
            & =  \sum \limits_{x \in \mathcal{X}} V^*(s_x) \underbrace{\sum \limits_{s^{\prime} \in x} 
        (\mathbb{P}(s^{\prime}|\: p_2) -   \mathbb{P}(s^{\prime}|\: p_1))}_{=0} \\
        & + \sum \limits_{x \in \mathcal{X}}\sum \limits_{s^{\prime} \in x} 
        (\mathbb{P}(s^{\prime}|\: p_1) -   \mathbb{P}(s^{\prime}|\: p_2)) \cdot d_{\text{s}}^{\prime}(s^{\prime},s_x) \\
        & = d_{\text{a}}(p_1,p_2) + R(p_1) - R(p_2).
    \end{aligned}
\end{equation}
Hence $d^*(p_1,p_2) = d_{\text{a}}(p_1,p_2)$.
Note that this proof holds for any choice of the abstract nodes' representatives. Lastly, let $(s_1,s_2) \in \mathcal{E}$ and $d_{\text{s}}(s_1,s_2) = d$. First note that if \mbox{$a \in \text{arg}\max\limits_{a^{\prime} \in \mathbb{A}(s_1)} Q^*(s_1,a^{\prime})$} and $((s_1,a),(s_2,\hat{a})) \in \mathcal{H}^{\prime}$ with $d^*_{\text{a}}((s_1,a),(s_2,\hat{a})) = d$ for some $\hat{a} \in \mathbb{A}(s_2)$ then \mbox{$\hat{a} \in \text{arg}\max\limits_{a^{\prime} \in \mathbb{A}(s_2)} Q^*(s_2,a^{\prime})$} because for all $b \in \mathbb{A}(s_2)$ it holds that
\begin{equation}
    \begin{aligned}
         Q^*(s_2,\hat{a}) = & \ Q^*(s_1,a ) - d^*_{\text{a}}((s_2,\hat{a}),(s_1,a)) 
        &  \\
        \geq &\  Q^*(s_1,\hat{b})  - d^*_{\text{a}}((s_2,\hat{a}),(s_1,a)) \\
        = & Q^*(s_2,b) - d^*_{\text{a}}((s_2,\hat{b}),(s_1,b))  - d^*_{\text{a}}((s_2,\hat{a}),(s_1,a)). \\
        = & Q^*(s_2,b) -(-d) + d = Q^*(s_2,b),
    \end{aligned}
\end{equation}
where $((s_1,\hat{b}),(s_2,b)) \in \mathcal{H}^{\prime}$ with $d^*_{\text{a}}((s_1,\hat{b}),(s_2,b)) = d$.
And since \mbox{$V^*(s) = \max\limits_{a \in \mathbb{A}(s)} Q^*(s,a)$} it directly follows that \mbox{$d = d_{\text{s}}(s_1,s_2) = d^*_{\text{s}}(s_1,s_2)$} \qed. 

\subsection{Runtime measurements}
Tab.~\ref{tab:runtimes} lists the average decision-making times for each environment of KVDA-UCT compared to OGA-UCT for 100 and 2000 iterations on states sampled from a distribution induced by random walks. The numbers show a median runtime overhead of $< 1\%$ for 100 iterations and $\approx 1\%$ for 2000 iterations.

\begin{table}[H]
\centering

\caption{Average decision-making times in milliseconds of KVDA-UCT compared to OGA-UCT 
This data was obtained using an Intel(R) Core(TM) i5-9600K CPU @ 3.70GHz.}
\label{tab:runtimes}
\scalebox{1.0}{
\begin{tabular}{l c c c c}
\hline
 Domain & KVDA-100 & OGA-100 & KVDA-2000 & OGA-2000 \\\hline

Academic Advising & 3.85 & 3.82 & 133.82 & 131.74\\

Cooperative Recon & 5.26 & 5.23 & 221.98 & 220.19\\

Earth Observation & 7.54 & 7.54 & 197.38 & 199.13\\

Elevators & 6.85 & 6.79 & 231.93 & 227.69\\

Game of Life & 4.99 & 4.96 & 139.38 & 134.06\\

Manufacturer & 10.26 & 11.01 & 281.03 & 294.91\\

Red Finned Blue Eye & 6.69 & 6.66 & 152.46 & 151.56\\

Sailing Wind & 4.11 & 4.09 & 139.39 & 140.82\\

Saving & 3.75 & 3.71 & 124.22 & 111.13\\

Skills Teaching & 4.97 & 4.94 & 219.90 & 216.68\\

SysAdmin  & 6.27 & 6.20 & 150.19 & 152.72\\

Tamarisk & 4.65 & 4.64 & 167.23 & 148.03\\

Traffic & 5.44 & 5.42 & 161.33 & 159.35\\

Push Your Luck & 5.84 & 5.78 & 170.25 & 175.19\\

Wildfire & 7.46 & 7.43 & 368.66 & 476.95\\

Wildlife Preserve & 7.06 & 7.05 & 2020.35 & 1685.53\\

Othello & 18.63 & 18.67 & 434.88 & 428.34\\

Connect4 & 4.40 & 4.33 & 133.35 & 123.58\\

Constrictor & 23.23 & 23.10 & 616.38 & 611.72\\

\bottomrule
\end{tabular}
}
\end{table}

\subsection[Domain-specific $\varepsilon_{a}$ values]{Domain-specific $\varepsilon_{a}$ values}
One problem with $(\varepsilon_{\text{a}},\varepsilon_{\text{t}})$-OGA is that while $\varepsilon_{\text{t}}$ is neatly bounded by $0$ and $2$, the $\varepsilon_{\text{a}}$ value has to be chosen on a per-environment basis since for example the value $\varepsilon_{\text{a}} = 0.5$ would be equivalent to $\varepsilon_{\text{a}} = 0$ for any environment with rewards that are all greater than $0.5$. Tab.~\ref{tab:epsa_values} lists the hand picked values for each environment that we chose for the experiment section. The values were chosen to be at the boundary of rewards that actually occur in these environments.

\begin{table}[h!]
\caption{A list of the environment-specific $\varepsilon_{\text{a}}$ values that were evaluated in the experiment section for $(\varepsilon_{\text{a}},0)$-OGA. All domains that are not explicitly listed here, used the default values $\varepsilon_{\text{a}} \in \{1,2,\infty\}$.}
\centering
\scalebox{0.8}{
\label{tab:epsa_values}
\centering
\begin{tabular}{l l}
\hline
\textbf{Environment} & \boldmath{$\varepsilon_{\text{a}}$ values} \\
\hline
Academic Advising & $\infty$ \\
Wildlife Preserve & 5, 10, 20, $\infty$ \\
Red Finned Blue Eye & 50, 100, 200, $\infty$ \\
Wildfire & 5, 10, 100, $\infty$ \\
Push Your Luck & 2, 5, 10, $\infty$ \\
Skill Teaching & 2, 3, $\infty$ \\
Tamarisk & 0.5, 1.0, $\infty$ \\
Cooperative Recon & 0.5, 1.0, $\infty$ \\
Manufacturer & 10, 20, $\infty$ \\
Connect 4 & 1, 5, 10, $\infty$ \\
Othello & 5, 10, 20, $\infty$ \\
Constrictor & 10, 20, 30, $\infty$ \\
Default & 1, 2, $\infty$ \\
\hline
\end{tabular}}
\end{table}

\subsection{Performances of $\varepsilon_{t}$-KVDA}

\begin{table}[H]
\centering

\caption{The parameter-optimized performances of $\varepsilon_{\text{t}}$-KVDA (our method) and $(\varepsilon_{\text{a}},\varepsilon_{\text{t}})$-OGA on the stochastic versions of the here-presented environments using 1000 MCTS iterations (left) and 100 MCTS iterations (right). The environments Push Your Luck and Wildlife Preserve from the deterministic experiments were excluded as computing their transition probabilities which is a requirement for both $\varepsilon_{\text{t}}$-KVDA and $(\varepsilon_{\text{a}},\varepsilon_{\text{t}})$-OGA is infeasible. Furthermore, Red Finned Blue Eye, Wildfire and Elevators were also excluded as the performances' variances were so high in these settings such that obtaining low confidence bounds was infeasible. Contrary to the stochastic setting, $\varepsilon_{\text{t}}$-KVDA does not consistently perform better than OGA. For example, while $\varepsilon_{\text{t}}$-KVDA can gain a clear advantage in Manufacturer and Sailing Wind, $(\varepsilon_{\text{a}},\varepsilon_{\text{t}})$-OGA is significantly better in Tamarisk.}
\label{tab:stochastic}
\begin{minipage}{0.45\textwidth}
\caption*{Average returns $(\uparrow)$ for 1000 MCTS iterations}
\scalebox{0.7}{
\setlength{\tabcolsep}{1mm}\begin{tabular}{ l c c }
\toprule
\textbf{Domain} & $(\varepsilon_{\text{a}},\varepsilon_{\text{t}})$-OGA & $\boldsymbol{\varepsilon_{\text{t}}}$\textbf{-KVDA} (ours)\\
\midrule
s-Academic Advising & $-63.9 \pm 0.8$ & $\boldsymbol { -63.9 \pm 0.8 }$\\
s-Cooperative Recon & $\boldsymbol { 14.3 \pm 0.2 }$ & $14.1 \pm 0.2$\\
s-Earth Observation & $-7.97 \pm 0.22$ & $\boldsymbol { -7.95 \pm 0.23 }$\\
s-Game of Life & $572.9 \pm 2.3$ & $\boldsymbol { 573.4 \pm 2.2 }$\\
s-Manufacturer & $-1141.0 \pm 11.1$ & $\boldsymbol { -863.3 \pm 10.7 }$\\
s-Sailing Wind & $-62.1 \pm 1.3$ & $\boldsymbol { -61.8 \pm 1.3 }$\\
s-Saving & $\boldsymbol { 50.7 \pm 0.1 }$ & $50.6 \pm 0.1$\\
s-Skills Teaching & $71.1 \pm 7.8$ & $\boldsymbol { 74.9 \pm 6.9 }$\\
s-SysAdmin & $\boldsymbol { 403.3 \pm 2.0 }$ & $403.1 \pm 2.1$\\
s-Tamarisk & $\boldsymbol { -532.9 \pm 8.2 }$ & $-563.6 \pm 8.3$\\
s-Traffic & $\boldsymbol { -13.4 \pm 0.3 }$ & $-13.6 \pm 0.3$\\
\bottomrule
\end{tabular}}
\end{minipage}
\hfill
\begin{minipage}{0.45\textwidth}
\centering
\caption*{Average returns $(\uparrow)$ for 100 MCTS iterations}
\scalebox{0.7}{
\setlength{\tabcolsep}{1mm}\begin{tabular}{ l c c }
\toprule
\textbf{Domain} & $(\varepsilon_{\text{a}},\varepsilon_{\text{t}})$-OGA & $\boldsymbol{\varepsilon_{\text{t}}}$\textbf{-KVDA} (ours)\\
\midrule
s-Academic Advising & $-88.0 \pm 1.2$ & $\boldsymbol { -87.8 \pm 1.2 }$\\
s-Cooperative Recon & $\boldsymbol { 7.08 \pm 0.37 }$ & $7.07 \pm 0.35$\\
s-Earth Observation & $\boldsymbol { -13.7 \pm 0.3 }$ & $-13.8 \pm 0.3$\\
s-Game of Life & $\boldsymbol { 498.7 \pm 3.2 }$ & $497.6 \pm 3.0$\\
s-Manufacturer & $-1457.6 \pm 21.5$ & $\boldsymbol { -1159.6 \pm 24.8 }$\\
s-Sailing Wind & $-78.3 \pm 1.1$ & $\boldsymbol { -73.2 \pm 1.2 }$\\
s-Saving & $\boldsymbol { 44.6 \pm 0.2 }$ & $44.5 \pm 0.2$\\
s-Skills Teaching & $\boldsymbol { -8.94 \pm 8.37 }$ & $-9.60 \pm 8.23$\\
s-SysAdmin & $\boldsymbol { 332.7 \pm 2.8 }$ & $331.1 \pm 2.7$\\
s-Tamarisk & $\boldsymbol { -767.9 \pm 9.8 }$ & $-836.1 \pm 7.8$\\
s-Traffic & $\boldsymbol { -22.0 \pm 0.4 }$ & $\boldsymbol { -22.0 \pm 0.4 }$\\

\bottomrule
\end{tabular}}
\end{minipage}
\end{table}

\subsection{Performances of parameter-optimized KVDA-UCT for 200 and 500 iterations}

\begin{table}[H]
\centering

\caption{Average returns $(\uparrow)$ using 500 MCTS iterations for KVDA-UCT, $(\infty,0)$-OGA, $(0,0)$-OGA (i.e. OGA-UCT), and the maximal performance of $(\varepsilon_{\text{a}},0)$-OGA when varying $\varepsilon_{\text{a}} > 0$. In the supplementary materials in Tab.\ref{tab:epsa_values} the domain-specific $\varepsilon_{\text{a}}$-values are listed. All performances are the maximal performances obtained by varying the exploration constants $C \in \{0.5,1,2,4,8,16\}$. Note that KVDA-UCT (our method) outperforms OGA-UCT in most environments. Furthermore, KVDA-UCT is either equal or performs better than parameter-optimized $(\varepsilon_{\text{a}},0)$-OGA even though KVDA-UCT introduces no extra parameter.}
\scalebox{0.8}{
\label{tab:param_optimized_others500}
\setlength{\tabcolsep}{1mm}\begin{tabular}{ l l l l l }
\toprule
\textbf{Domain} & $(0,0)$-OGA & $(\infty,0)$-OGA & $(\varepsilon_{\text{a}},0)$-OGA & \textbf{KVDA-UCT} (ours)\\
\midrule
Connect4 & $36.7 \pm 0.5$ & $\boldsymbol { 41.3 \pm 1.0 }$ & $\boldsymbol { 41.3 \pm 1.0 }$ & $39.9 \pm 0.6$\\
Constrictor & $\boldsymbol { 94.2 \pm 0.2 }$ & $92.8 \pm 0.2$ & $94.2 \pm 0.3$ & $94.0 \pm 0.2$\\
Othello & $\boldsymbol { 175.3 \pm 1.1 }$ & $156.8 \pm 0.8$ & $175.2 \pm 1.6$ & $173.4 \pm 1.1$\\
Pusher & $104.5 \pm 0.0$ & $104.5 \pm 0.0$ & $104.5 \pm 0.0$ & $\boldsymbol { 104.5 \pm 0.0 }$\\
d-Academic Advising & $-42.1 \pm 0.3$ & $\boldsymbol { -42.1 \pm 0.3 }$ & $\boldsymbol { -42.1 \pm 0.3 }$ & $-42.3 \pm 0.3$\\
d-Cooperative Recon & $15.3 \pm 0.1$ & $14.4 \pm 0.1$ & $\boldsymbol { 15.4 \pm 0.2 }$ & $15.3 \pm 0.2$\\
d-Earth Observation & $-7.28 \pm 0.12$ & $-29.6 \pm 0.4$ & $-29.6 \pm 0.4$ & $\boldsymbol { -6.98 \pm 0.10 }$\\
d-Elevators & $-15.6 \pm 0.3$ & $\boldsymbol { -14.6 \pm 0.3 }$ & $\boldsymbol { -14.6 \pm 0.3 }$ & $-14.9 \pm 0.3$\\
d-Game of Life & $659.9 \pm 2.6$ & $659.9 \pm 2.6$ & $660.7 \pm 2.5$ & $\boldsymbol { 661.6 \pm 2.5 }$\\
d-Manufacturer & $-1270.2 \pm 11.7$ & $-1646.1 \pm 20.8$ & $-1259.1 \pm 16.5$ & $\boldsymbol { -1178.3 \pm 22.3 }$\\
d-Push Your Luck & $124.2 \pm 1.9$ & $67.4 \pm 1.0$ & $131.4 \pm 2.0$ & $\boldsymbol { 136.5 \pm 1.9 }$\\
d-RedFinnedBlueEye & $7978.4 \pm 32.9$ & $7738.1 \pm 34.2$ & $7777.4 \pm 34.3$ & $\boldsymbol { 8041.7 \pm 33.1 }$\\
d-Sailing Wind & $-43.4 \pm 0.7$ & $-42.8 \pm 0.7$ & $-42.8 \pm 0.7$ & $\boldsymbol { -41.6 \pm 0.7 }$\\
d-Saving & $\boldsymbol { 64.5 \pm 0.2 }$ & $61.7 \pm 0.2$ & $64.2 \pm 0.2$ & $64.4 \pm 0.2$\\
d-Skills Teaching & $202.8 \pm 3.7$ & $201.0 \pm 3.9$ & $202.6 \pm 3.9$ & $\boldsymbol { 206.5 \pm 3.7 }$\\
d-SysAdmin & $477.7 \pm 1.5$ & $448.8 \pm 1.2$ & $450.8 \pm 1.2$ & $\boldsymbol { 478.1 \pm 1.5 }$\\
d-Tamarisk & $-228.9 \pm 4.1$ & $-217.9 \pm 2.4$ & $-217.9 \pm 2.4$ & $\boldsymbol { -197.2 \pm 2.7 }$\\
d-Traffic & $-2.60 \pm 0.10$ & $\boldsymbol { -2.53 \pm 0.09 }$ & $\boldsymbol { -2.53 \pm 0.09 }$ & $\boldsymbol { -2.53 \pm 0.09 }$\\
d-Wildfire & $\boldsymbol { -183.6 \pm 36.1 }$ & $-506.0 \pm 36.2$ & $-409.6 \pm 36.1$ & $-190.6 \pm 36.3$\\
d-Wildlife Preserve & $1385.1 \pm 0.8$ & $1384.2 \pm 0.8$ & $\boldsymbol { 1385.5 \pm 0.8 }$ & $1385.0 \pm 0.8$\\
\bottomrule
\end{tabular}}
\end{table}

\begin{table}[H]
\centering

\caption{Average returns using 200 $(\uparrow)$ MCTS iterations for KVDA-UCT, $(\infty,0)$-OGA, $(0,0)$-OGA (i.e. OGA-UCT), and the maximal performance of $(\varepsilon_{\text{a}},0)$-OGA when varying $\varepsilon_{\text{a}} > 0$. In the supplementary materials in Tab.\ref{tab:epsa_values} the domain-specific $\varepsilon_{\text{a}}$-values are listed. All performances are the maximal performances obtained by varying the exploration constants $C \in \{0.5,1,2,4,8,16\}$. Note that KVDA-UCT (our method) outperforms OGA-UCT in most environments. Furthermore, KVDA-UCT is either equal or performs better than parameter-optimized $(\varepsilon_{\text{a}},0)$-OGA even though KVDA-UCT introduces no extra parameter.}
\label{tab:param_optimized_others200}
\centering
\setlength{\tabcolsep}{1mm}\begin{tabular}{ l l l l l }
\toprule
\textbf{Domain} & $(0,0)$-OGA & $(\infty,0)$-OGA & $(\varepsilon_{\text{a}},0)$-OGA & \textbf{KVDA-UCT} (ours)\\
\midrule
Connect4 & $29.1 \pm 0.4$ & $\boldsymbol { 29.9 \pm 0.6 }$ & $\boldsymbol { 29.9 \pm 0.6 }$ & $29.5 \pm 0.4$\\
Constrictor & $91.3 \pm 0.3$ & $90.2 \pm 0.2$ & $\boldsymbol { 91.3 \pm 0.4 }$ & $91.2 \pm 0.3$\\
Othello & $164.5 \pm 1.2$ & $152.1 \pm 0.8$ & $\boldsymbol { 164.7 \pm 1.7 }$ & $164.5 \pm 1.2$\\
Pusher & $104.5 \pm 0.0$ & $\boldsymbol { 104.5 \pm 0.0 }$ & $\boldsymbol { 104.5 \pm 0.0 }$ & $104.5 \pm 0.0$\\
d-Academic Advising & $-47.3 \pm 0.4$ & $\boldsymbol { -47.2 \pm 0.4 }$ & $\boldsymbol { -47.2 \pm 0.4 }$ & $-47.5 \pm 0.4$\\
d-Cooperative Recon & $\boldsymbol { 13.2 \pm 0.3 }$ & $12.7 \pm 0.3$ & $13.1 \pm 0.3$ & $13.1 \pm 0.3$\\
d-Earth Observation & $-7.82 \pm 0.14$ & $-29.3 \pm 0.2$ & $-29.3 \pm 0.2$ & $\boldsymbol { -7.11 \pm 0.11 }$\\
d-Elevators & $-16.2 \pm 0.3$ & $-16.1 \pm 0.3$ & $-16.1 \pm 0.3$ & $\boldsymbol { -16.0 \pm 0.3 }$\\
d-Game of Life & $651.4 \pm 2.5$ & $\boldsymbol { 652.3 \pm 2.4 }$ & $\boldsymbol { 652.3 \pm 2.4 }$ & $651.1 \pm 2.5$\\
d-Manufacturer & $-1370.7 \pm 15.3$ & $-1641.4 \pm 21.2$ & $-1320.7 \pm 18.6$ & $\boldsymbol { -1180.8 \pm 22.2 }$\\
d-Push Your Luck & $122.6 \pm 1.9$ & $67.1 \pm 0.9$ & $\boldsymbol { 129.6 \pm 1.9 }$ & $125.9 \pm 1.8$\\
d-RedFinnedBlueEye & $7713.6 \pm 33.9$ & $7511.8 \pm 34.7$ & $7513.4 \pm 34.9$ & $\boldsymbol { 7716.9 \pm 33.7 }$\\
d-Sailing Wind & $-52.7 \pm 0.8$ & $-52.3 \pm 0.8$ & $-52.3 \pm 0.8$ & $\boldsymbol { -51.7 \pm 0.8 }$\\
d-Saving & $\boldsymbol { 60.9 \pm 0.2 }$ & $58.2 \pm 0.2$ & $60.5 \pm 0.2$ & $60.6 \pm 0.2$\\
d-Skills Teaching & $183.9 \pm 3.9$ & $\boldsymbol { 186.7 \pm 3.8 }$ & $\boldsymbol { 186.7 \pm 3.8 }$ & $185.0 \pm 3.8$\\
d-SysAdmin & $476.6 \pm 1.6$ & $450.2 \pm 1.2$ & $450.3 \pm 1.2$ & $\boldsymbol { 478.9 \pm 1.4 }$\\
d-Tamarisk & $-265.1 \pm 5.5$ & $-242.9 \pm 3.5$ & $-242.9 \pm 3.5$ & $\boldsymbol { -221.7 \pm 3.8 }$\\
d-Traffic & $\boldsymbol { -5.34 \pm 0.15 }$ & $\boldsymbol { -5.34 \pm 0.15 }$ & $\boldsymbol { -5.34 \pm 0.15 }$ & $\boldsymbol { -5.34 \pm 0.15 }$\\
d-Wildfire & $-193.8 \pm 36.1$ & $-476.5 \pm 36.1$ & $-432.2 \pm 36.2$ & $\boldsymbol { -191.2 \pm 36.1 }$\\
d-Wildlife Preserve & $\boldsymbol { 1390.1 \pm 0.9 }$ & $1389.4 \pm 0.9$ & $1389.4 \pm 0.9$ & $1389.5 \pm 1.0$\\
\bottomrule
\end{tabular}
\end{table}

\subsection{Performance graphs of parameter-optimized KVDA-UCT}
\label{sec:performance_graphs}

\begin{figure}[H]
\centering

\begin{minipage}{0.23\textwidth}
\centering
\includegraphics[width=\linewidth]{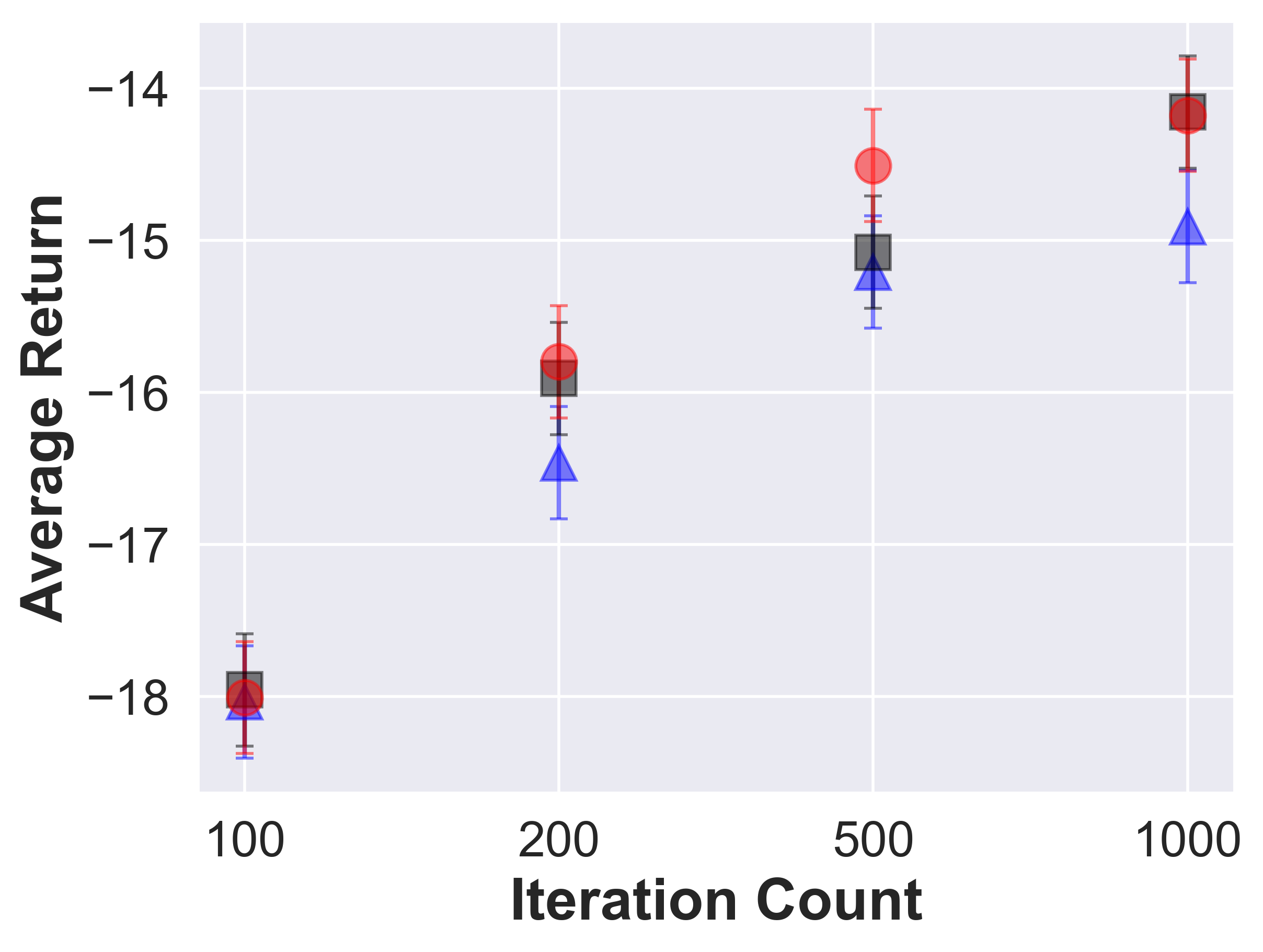}
\caption*{(a) Elevators}
\end{minipage}
\hfill
\begin{minipage}{0.23\textwidth}
\centering
\includegraphics[width=\linewidth]{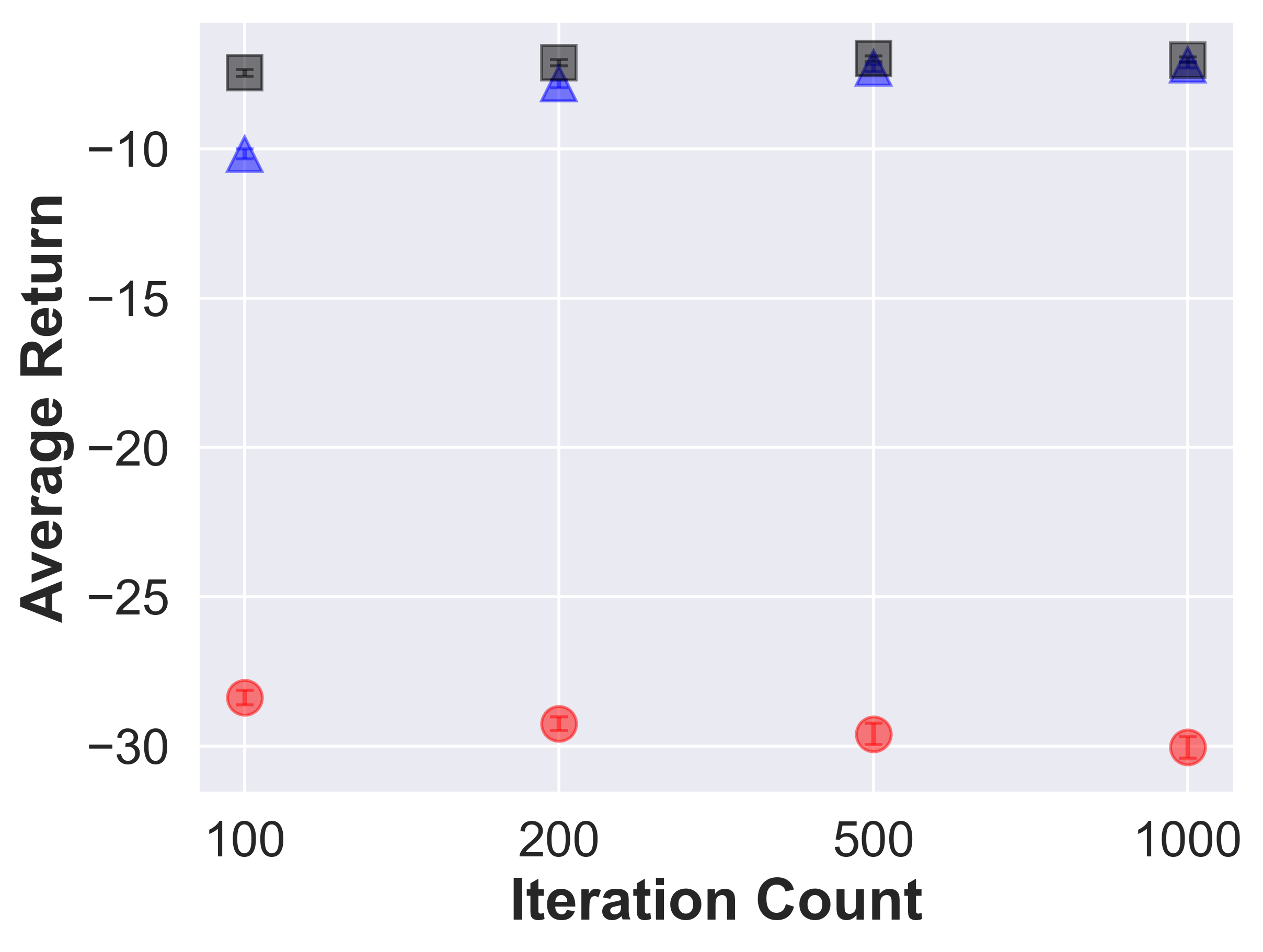}
\caption*{(b) Earth Observation}
\end{minipage}
\hfill
\begin{minipage}{0.23\textwidth}
\centering
\includegraphics[width=\linewidth]{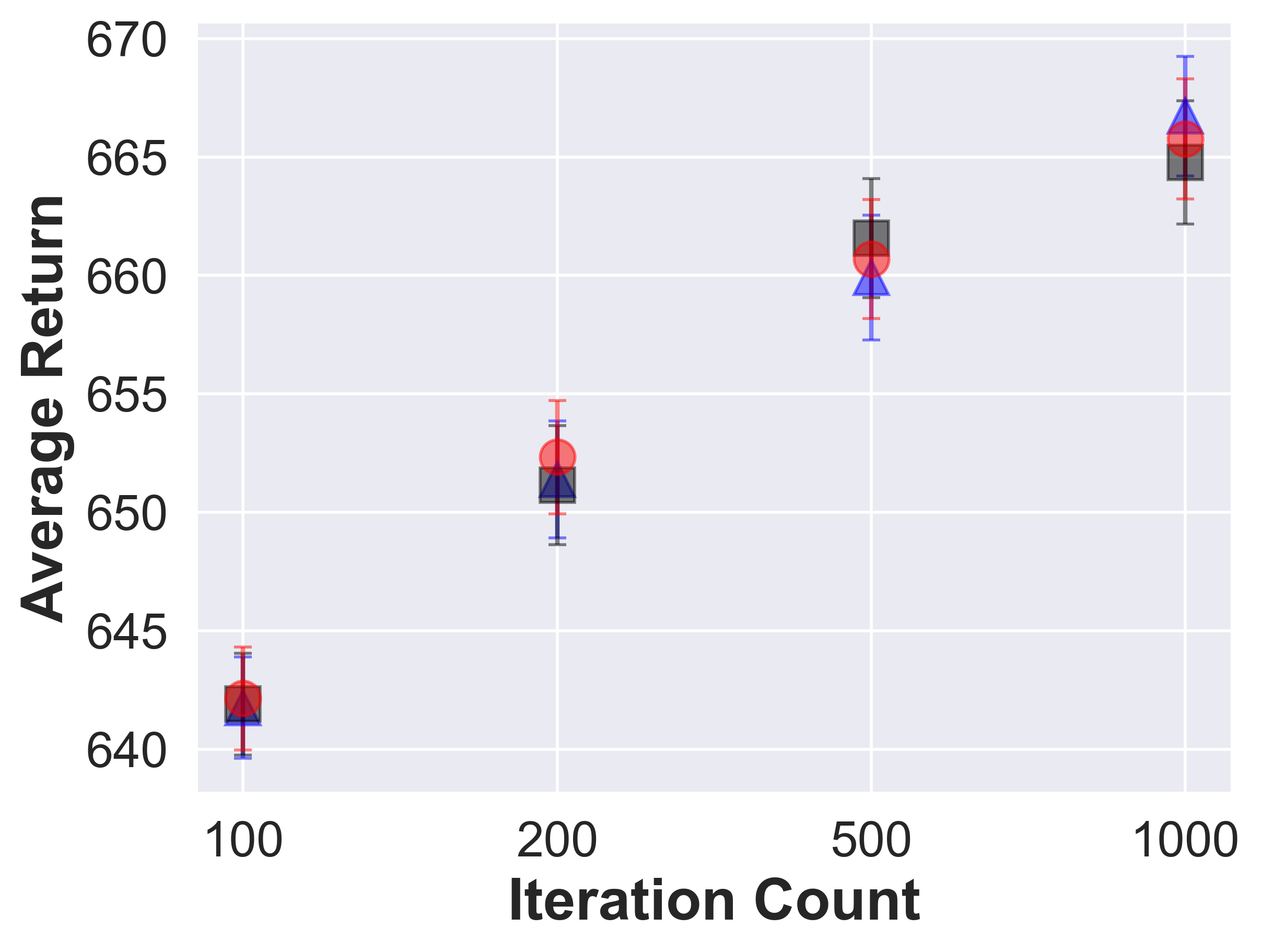}
\caption*{(c) Game of Life}
\end{minipage}
\hfill
\begin{minipage}{0.23\textwidth}
\centering
\includegraphics[width=\linewidth]{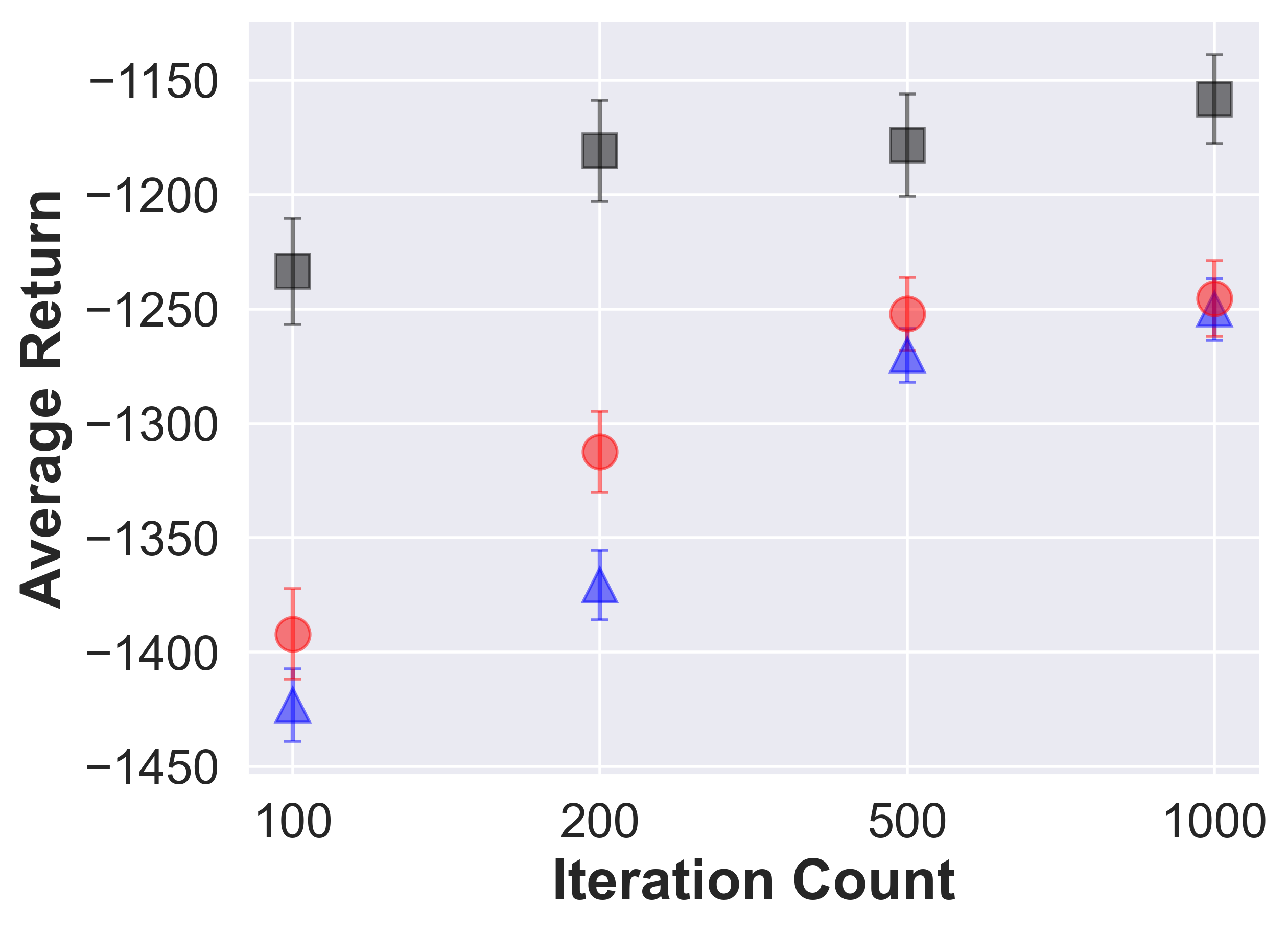}
\caption*{(d) Manufacturer}
\end{minipage}
\hfill
\begin{minipage}{0.23\textwidth}
\centering
\includegraphics[width=\linewidth]{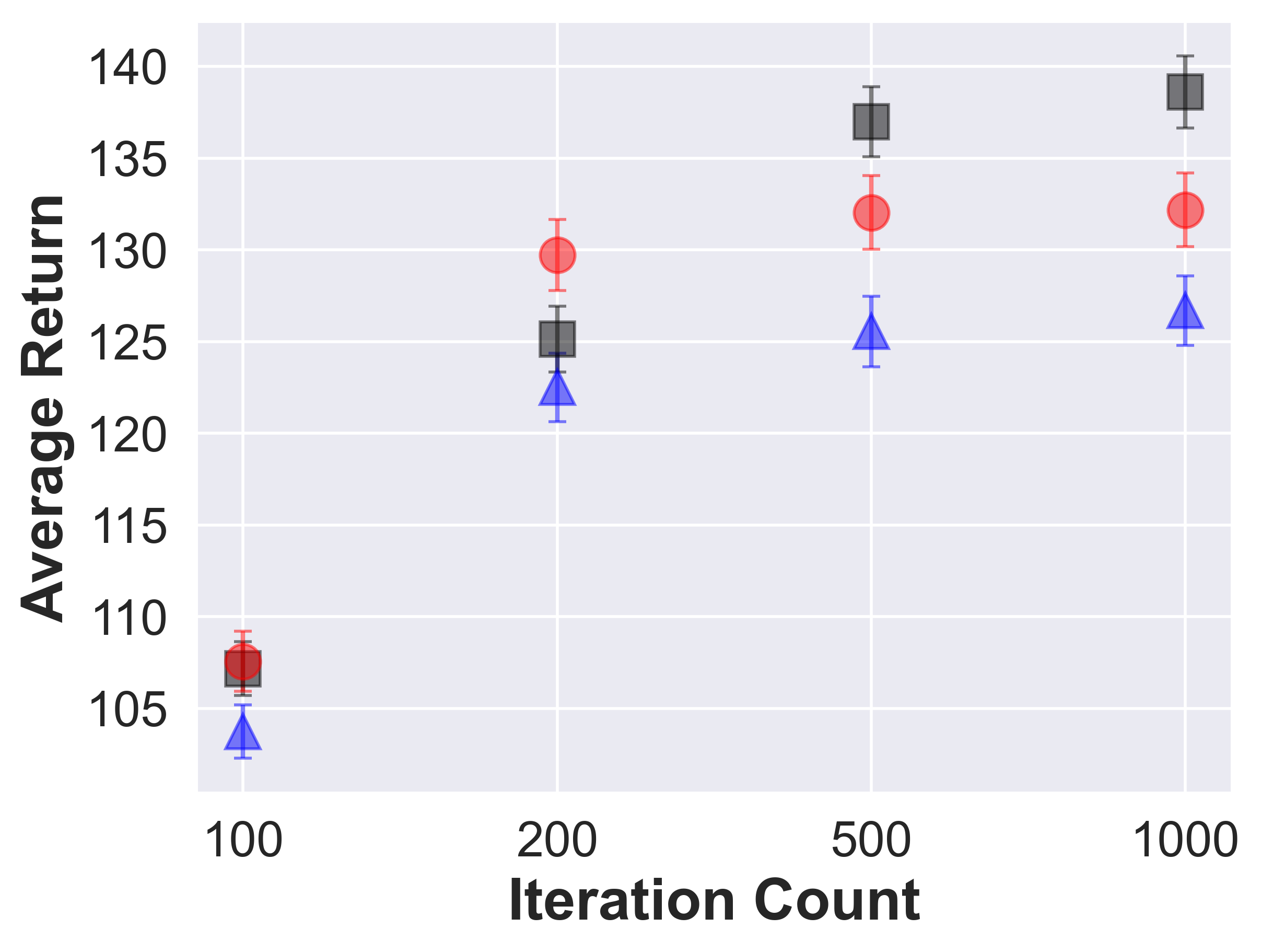}
\caption*{(e) Push Your Luck}
\end{minipage}
\hfill
\begin{minipage}{0.23\textwidth}
\centering
\includegraphics[width=\linewidth]{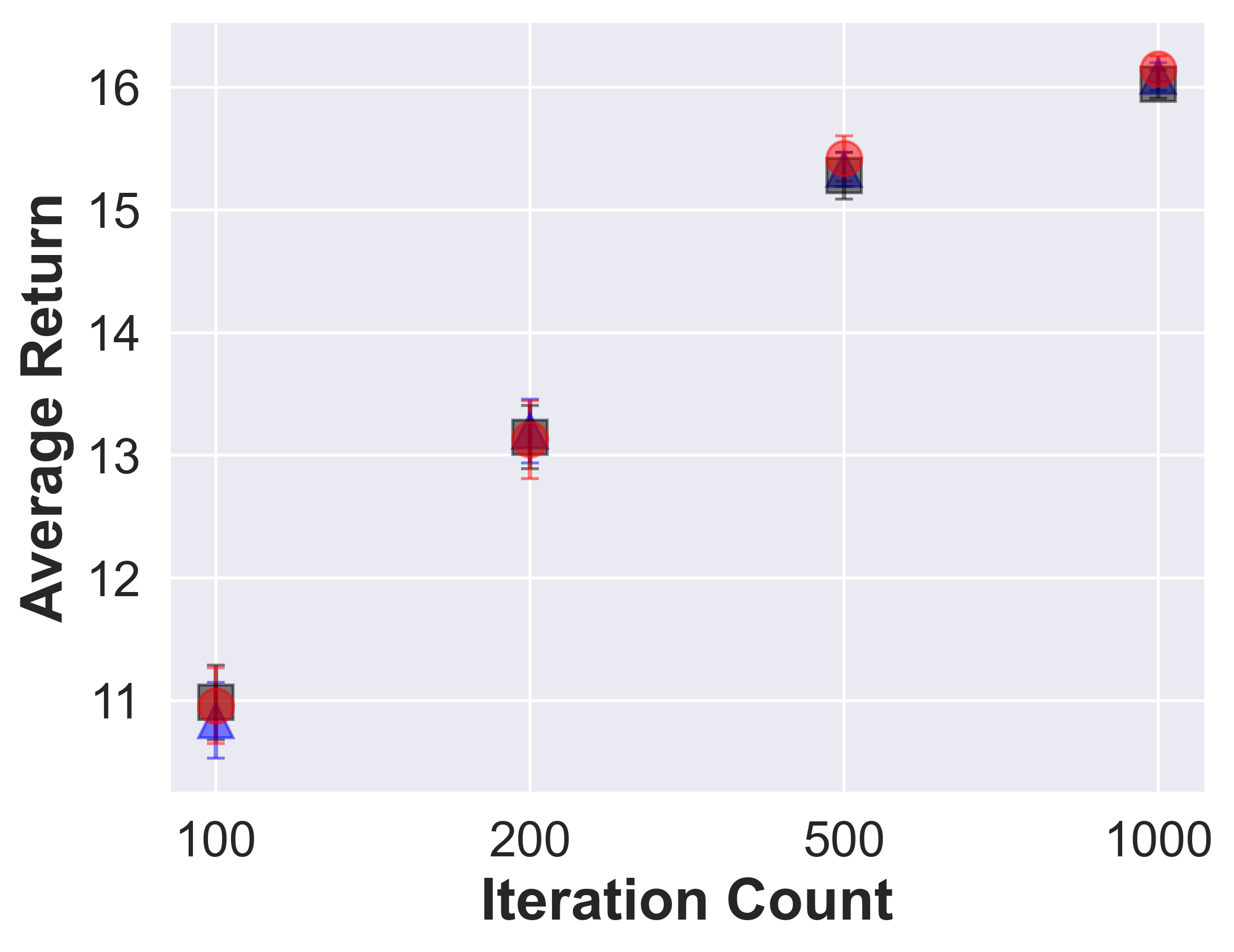}
\caption*{(f) Cooperative Recon}
\end{minipage}
\hfill
\begin{minipage}{0.23\textwidth}
\centering
\includegraphics[width=\linewidth]{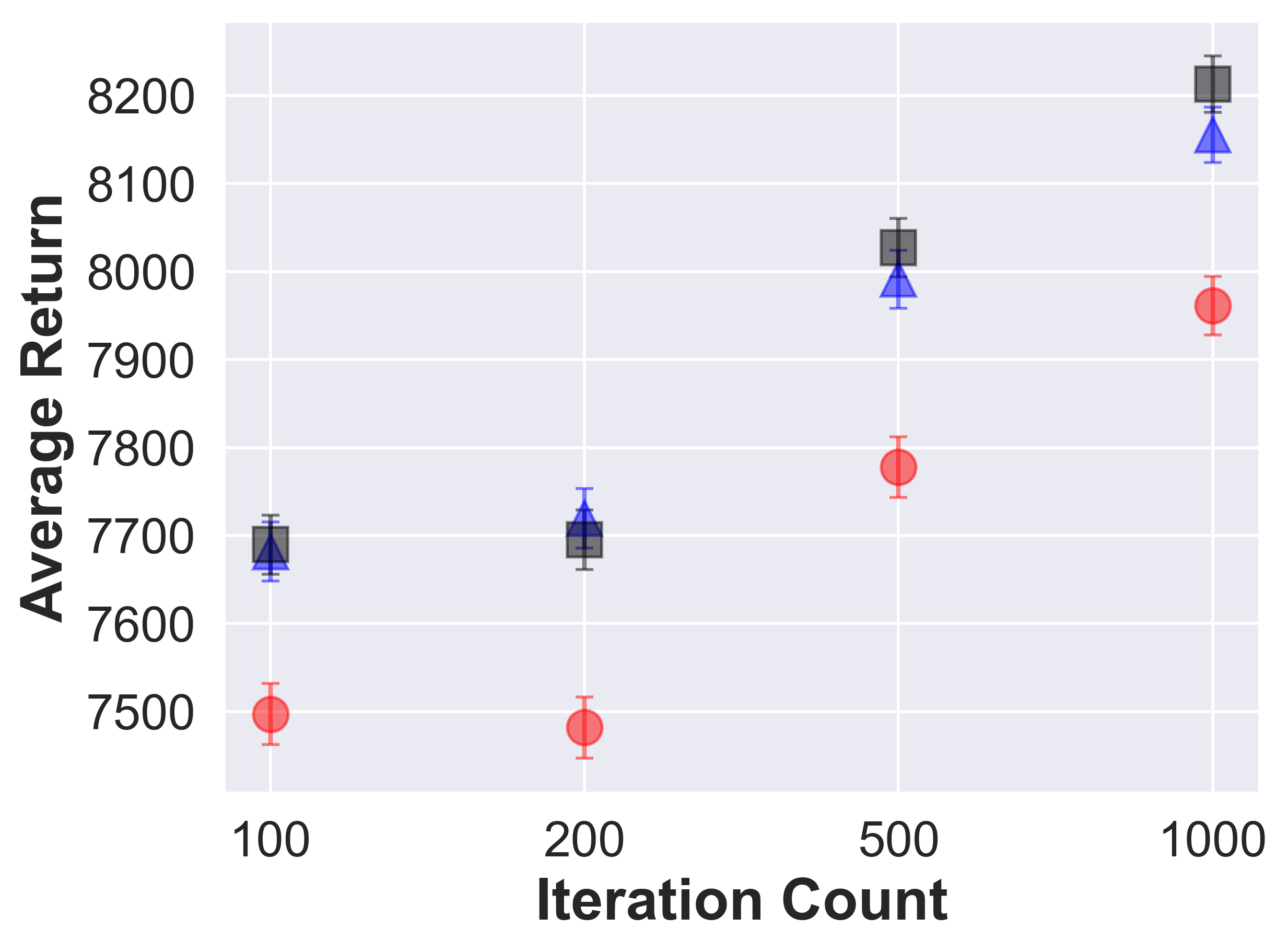}
\caption*{(g) Red Finned Blue Eye}
\end{minipage}
\hfill
\begin{minipage}{0.23\textwidth}
\centering
\includegraphics[width=\linewidth]{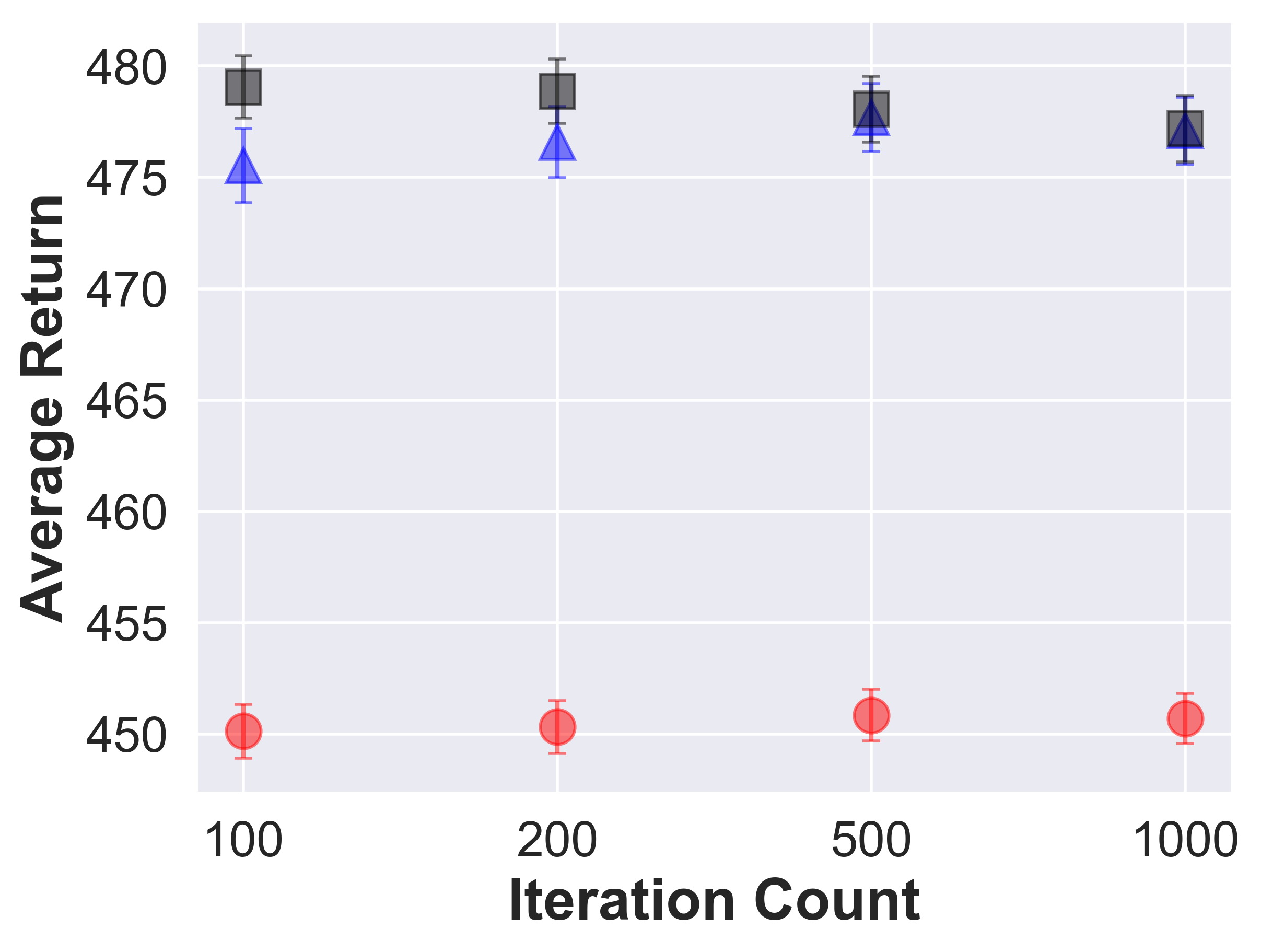}
\caption*{(h) SysAdmin}
\end{minipage}
\hfill
\begin{minipage}{0.23\textwidth}
\centering
\includegraphics[width=\linewidth]{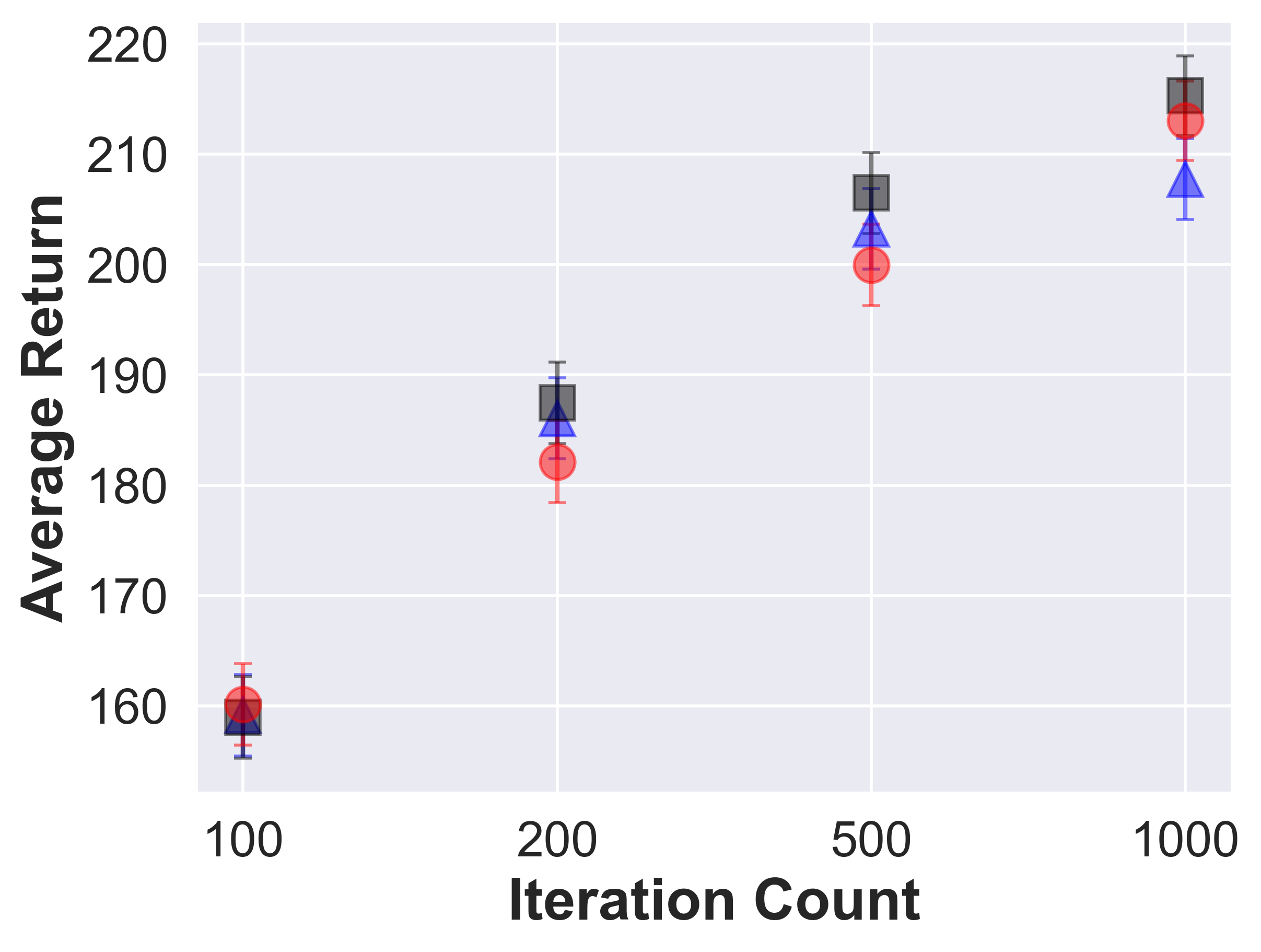}
\caption*{(i) Skill Teaching}
\end{minipage}
\hfill
\begin{minipage}{0.23\textwidth}
\centering
\includegraphics[width=\linewidth]{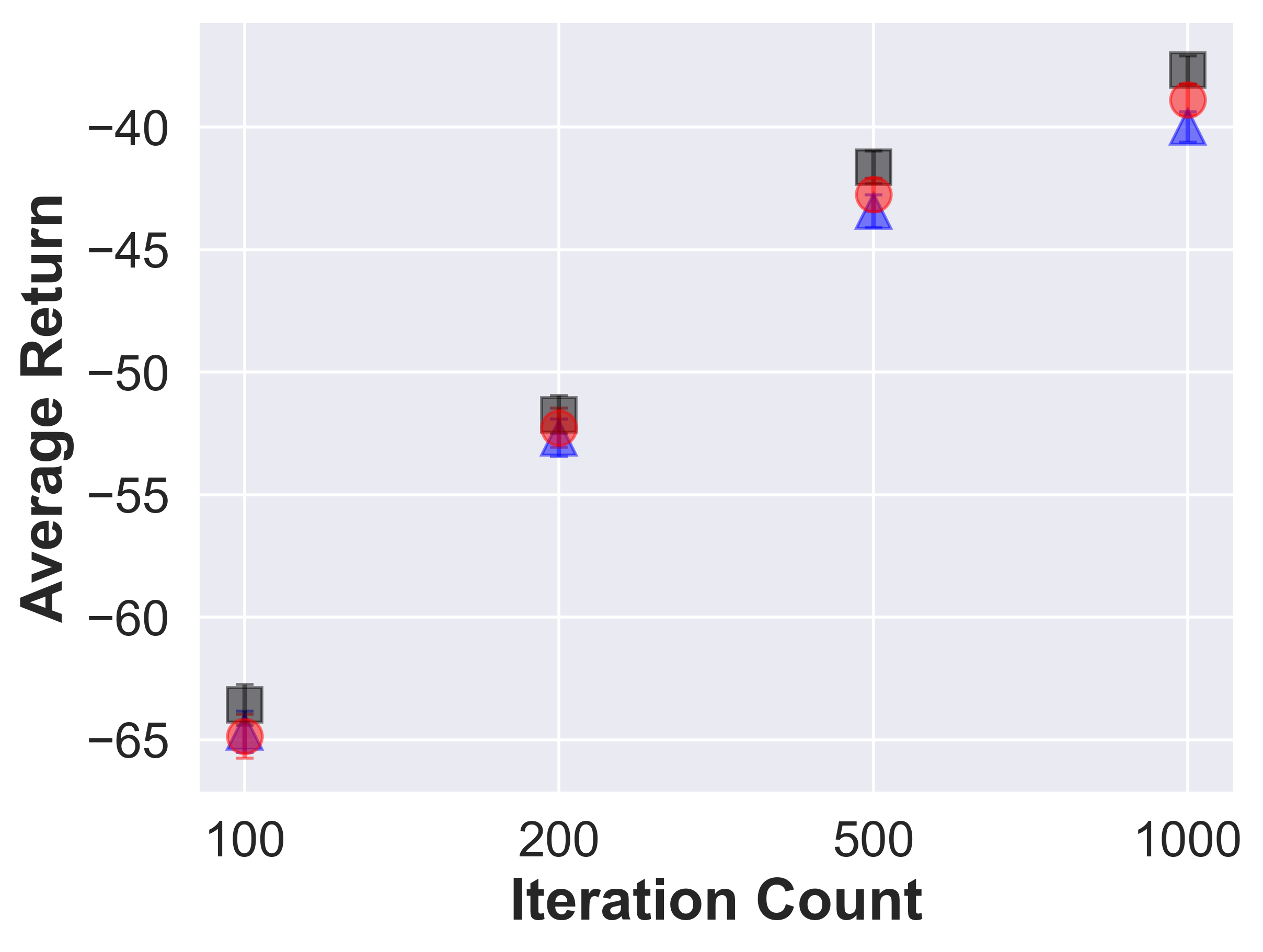}
\caption*{(j) Sailing Wind}
\end{minipage}
\hfill
\begin{minipage}{0.23\textwidth}
\centering
\includegraphics[width=\linewidth]{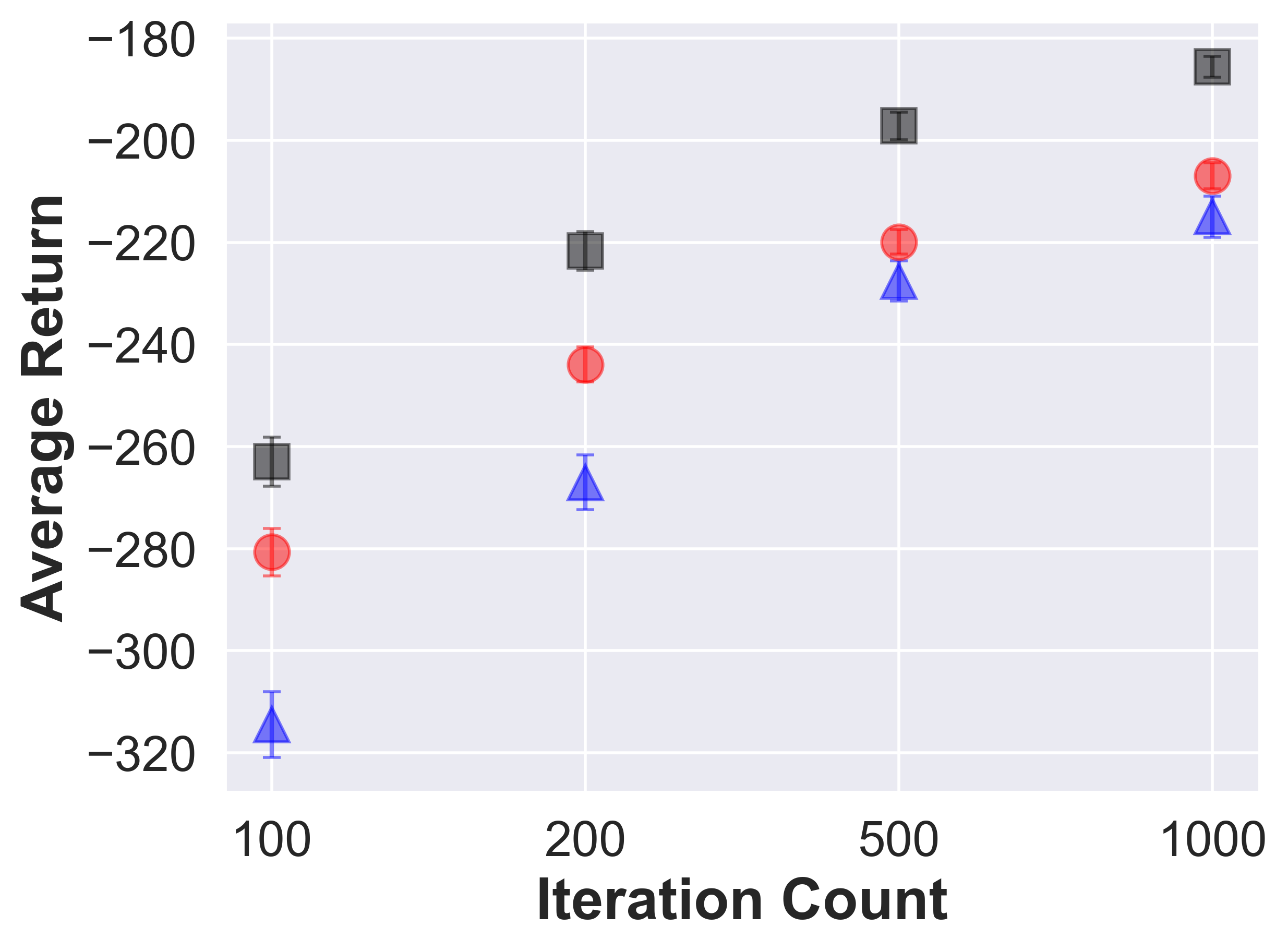}
\caption*{(k) Tamarisk}
\end{minipage}
\hfill
\begin{minipage}{0.23\textwidth}
\centering
\includegraphics[width=\linewidth]{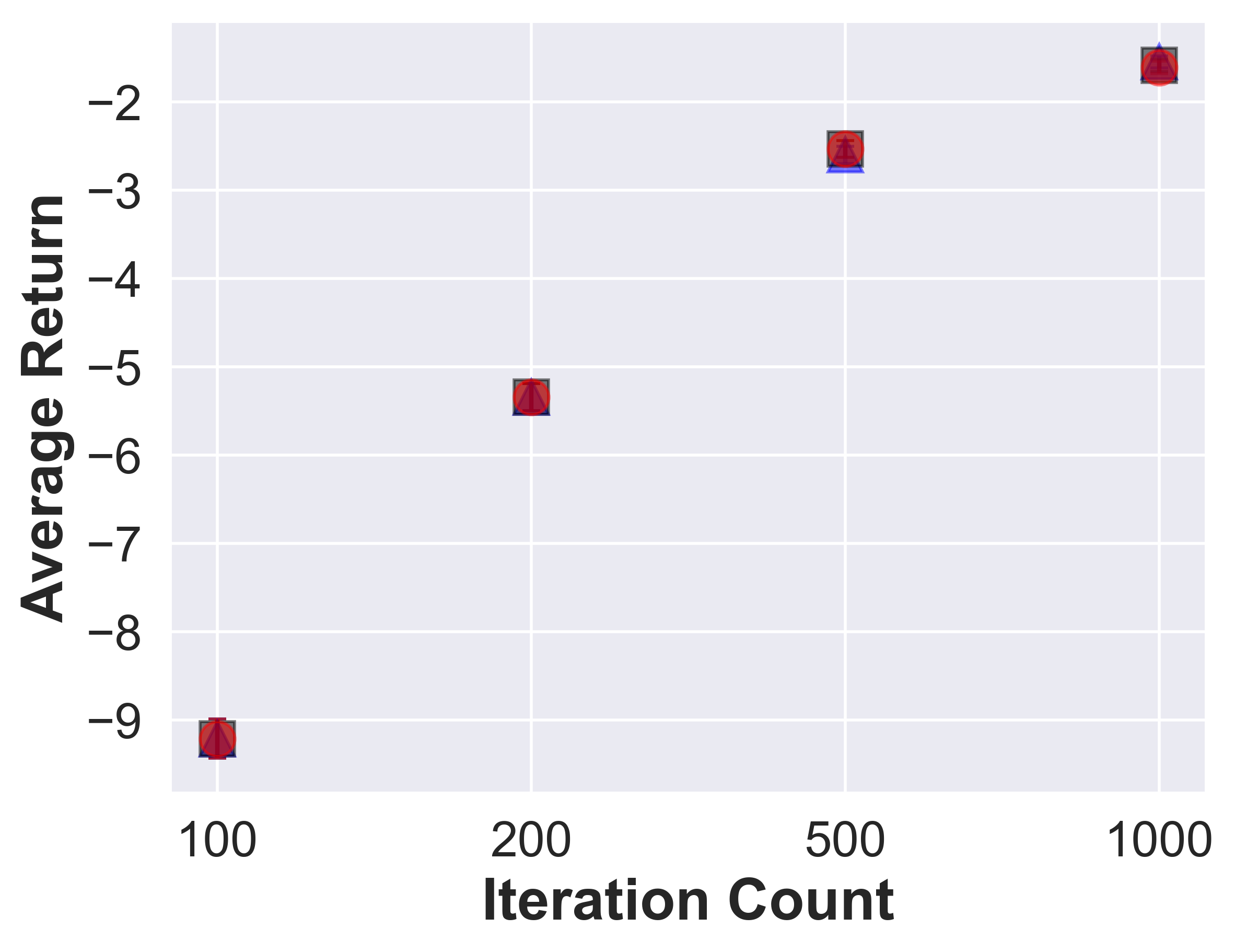}
\caption*{(l) Traffic}
\end{minipage}
\hfill
\begin{minipage}{0.23\textwidth}
\centering
\includegraphics[width=\linewidth]{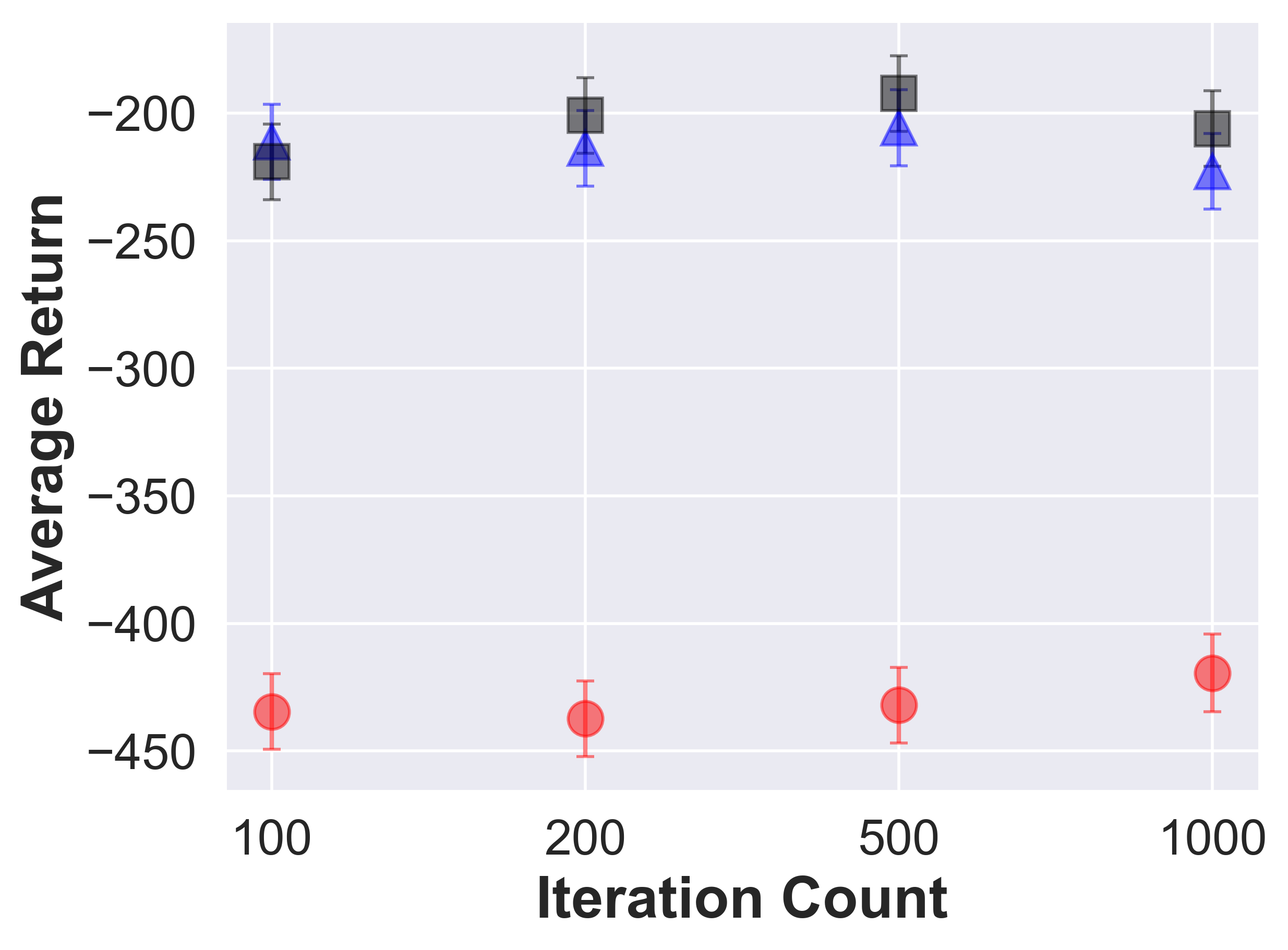}
\caption*{(m) Wildfire}
\end{minipage}
\hfill
\begin{minipage}{0.23\textwidth}
\centering
\includegraphics[width=\linewidth]{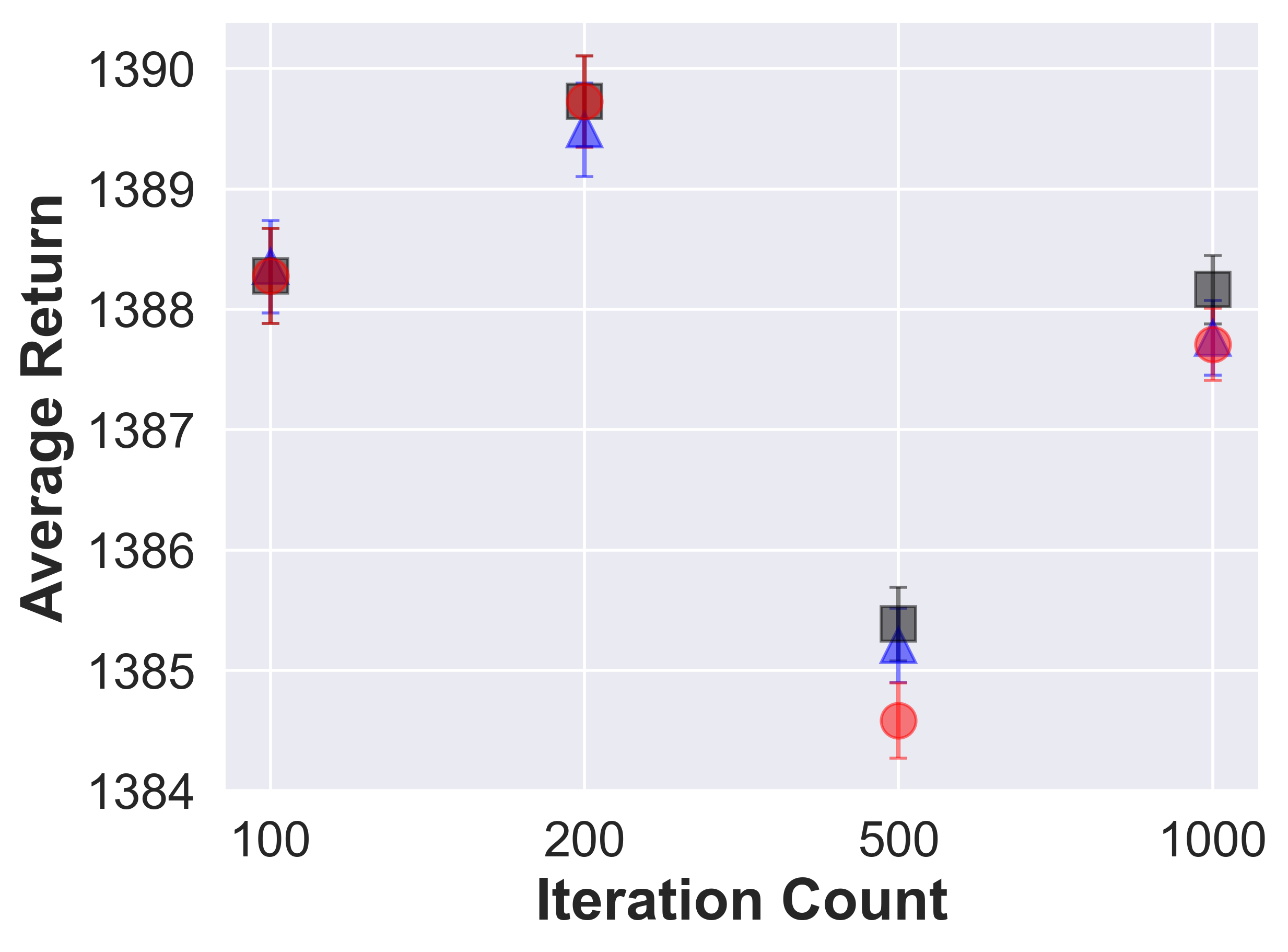}
\caption*{(n) Wildlife Preserve}
\end{minipage}
\hfill
\begin{minipage}{0.23\textwidth}
\centering
\includegraphics[width=\linewidth]{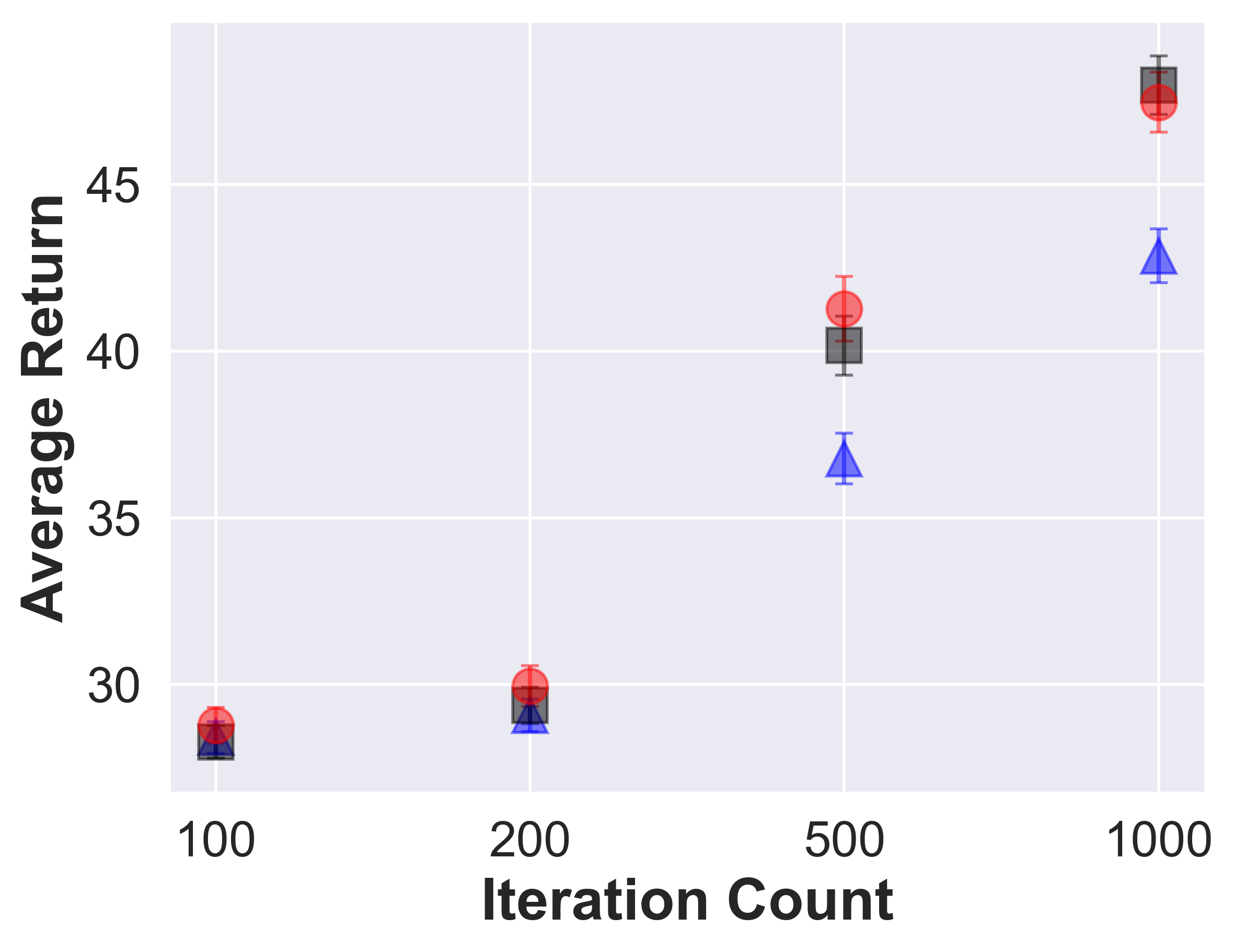}
\caption*{(o) Connect 4}
\end{minipage}
\hfill
\begin{minipage}{0.23\textwidth}
\centering
\includegraphics[width=\linewidth]{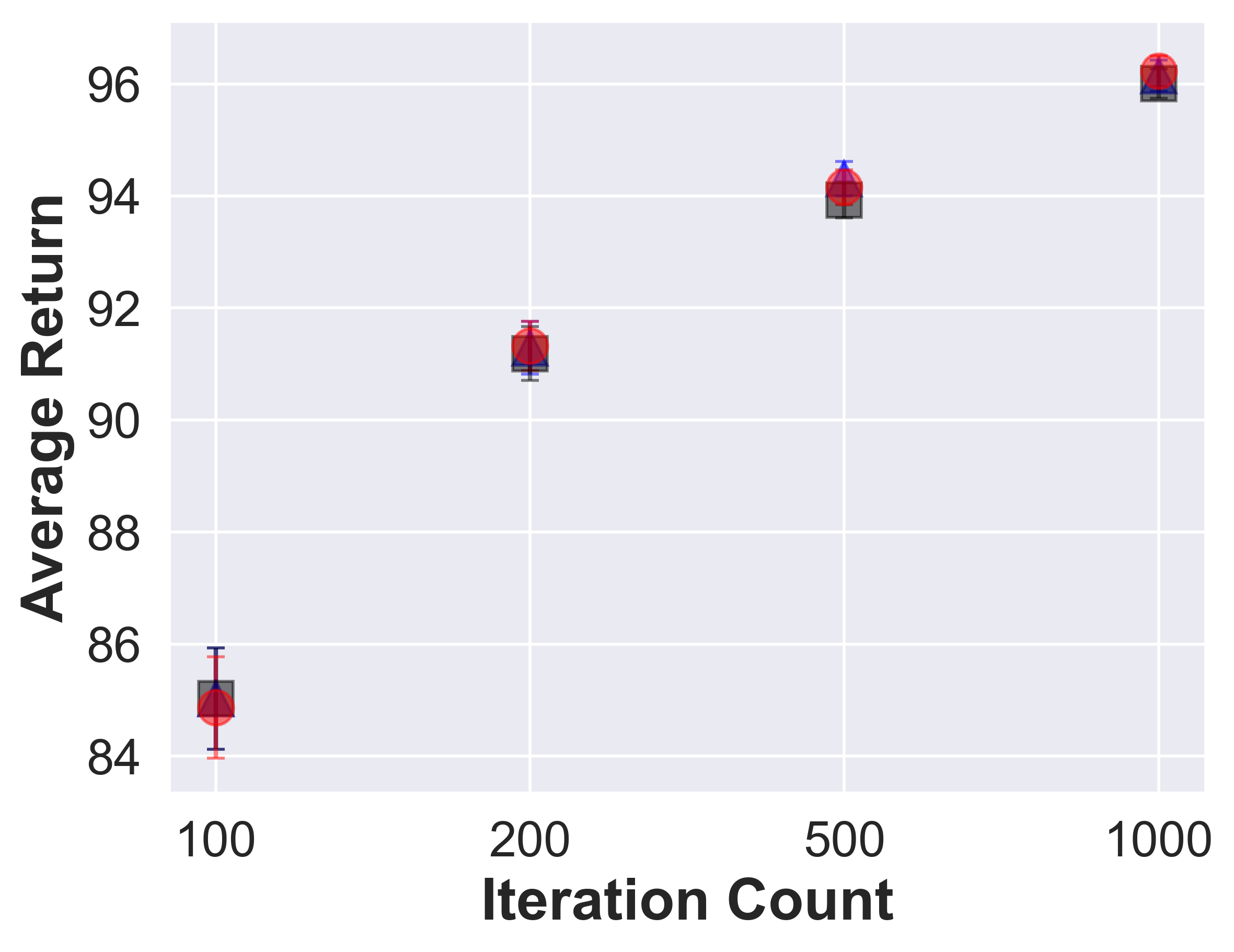}
\caption*{(p) Constrictor}
\end{minipage}
\hfill
\begin{minipage}{0.23\textwidth}
\centering
\includegraphics[width=\linewidth]{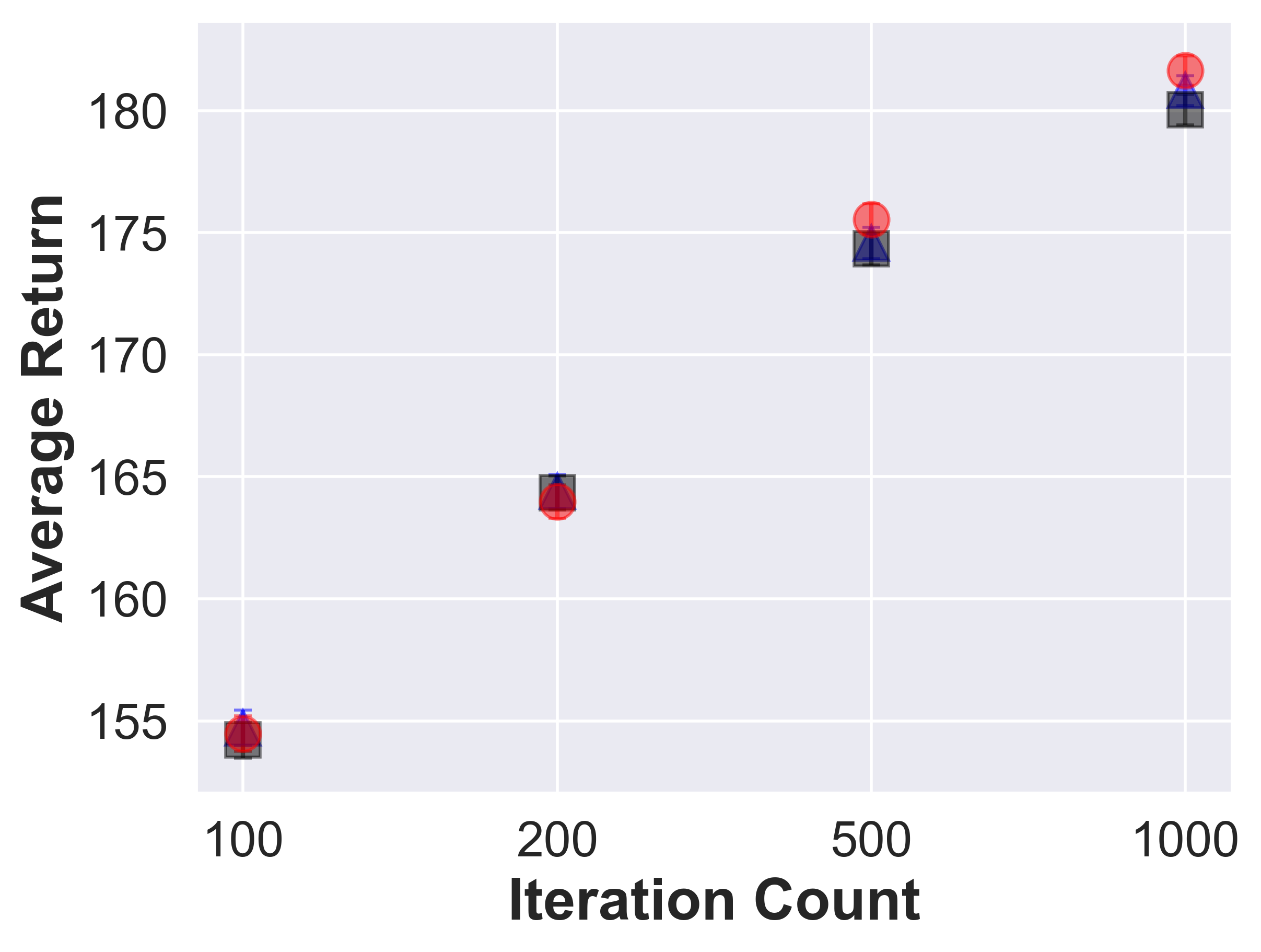}
\caption*{(q) Othello}
\end{minipage}
\hfill
\begin{minipage}{0.23\textwidth}
\centering
\includegraphics[width=\linewidth]{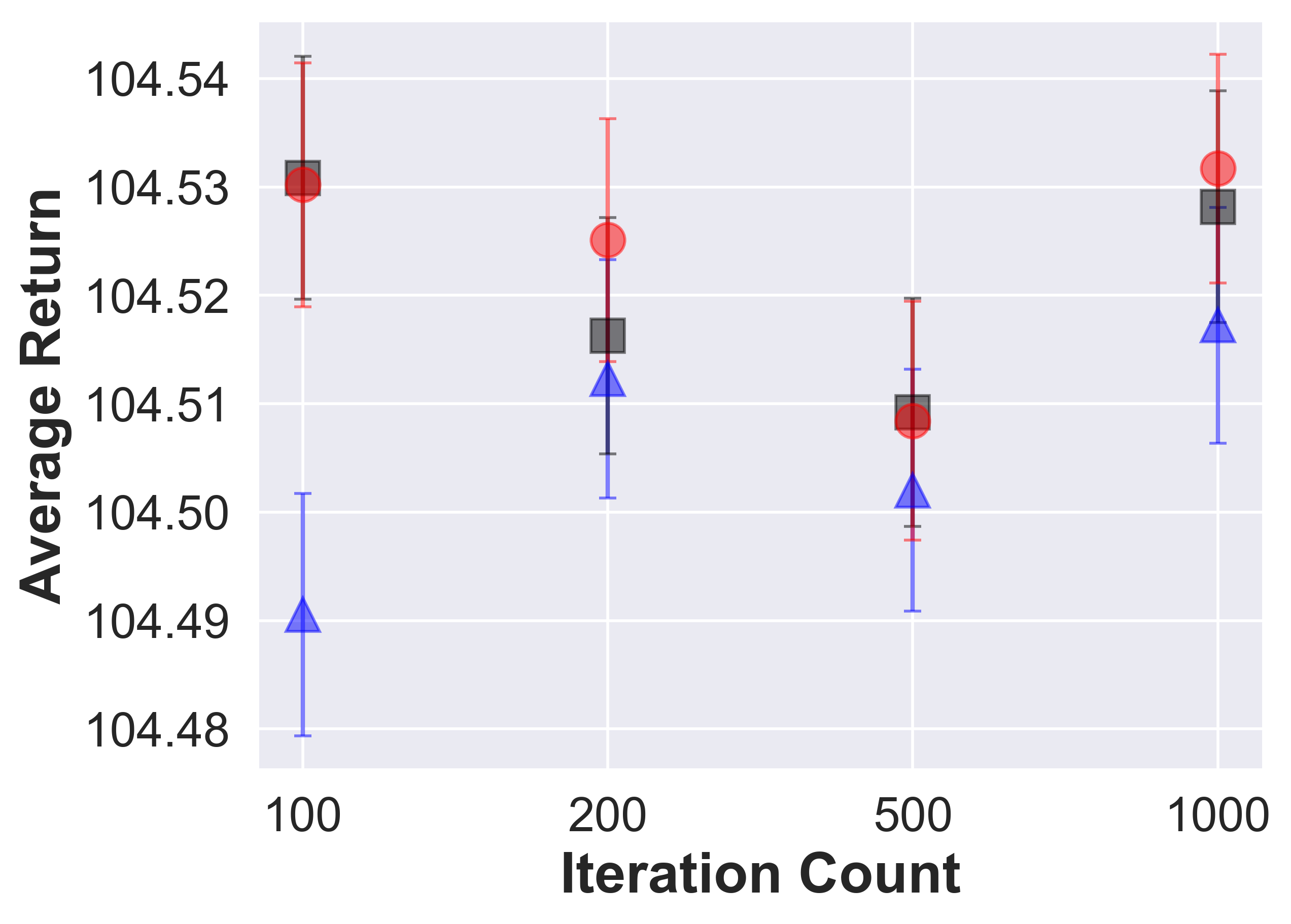}
\caption*{(r) Pusher}
\end{minipage}
\hfill
\begin{minipage}{0.23\textwidth}
\centering
\includegraphics[width=\linewidth]{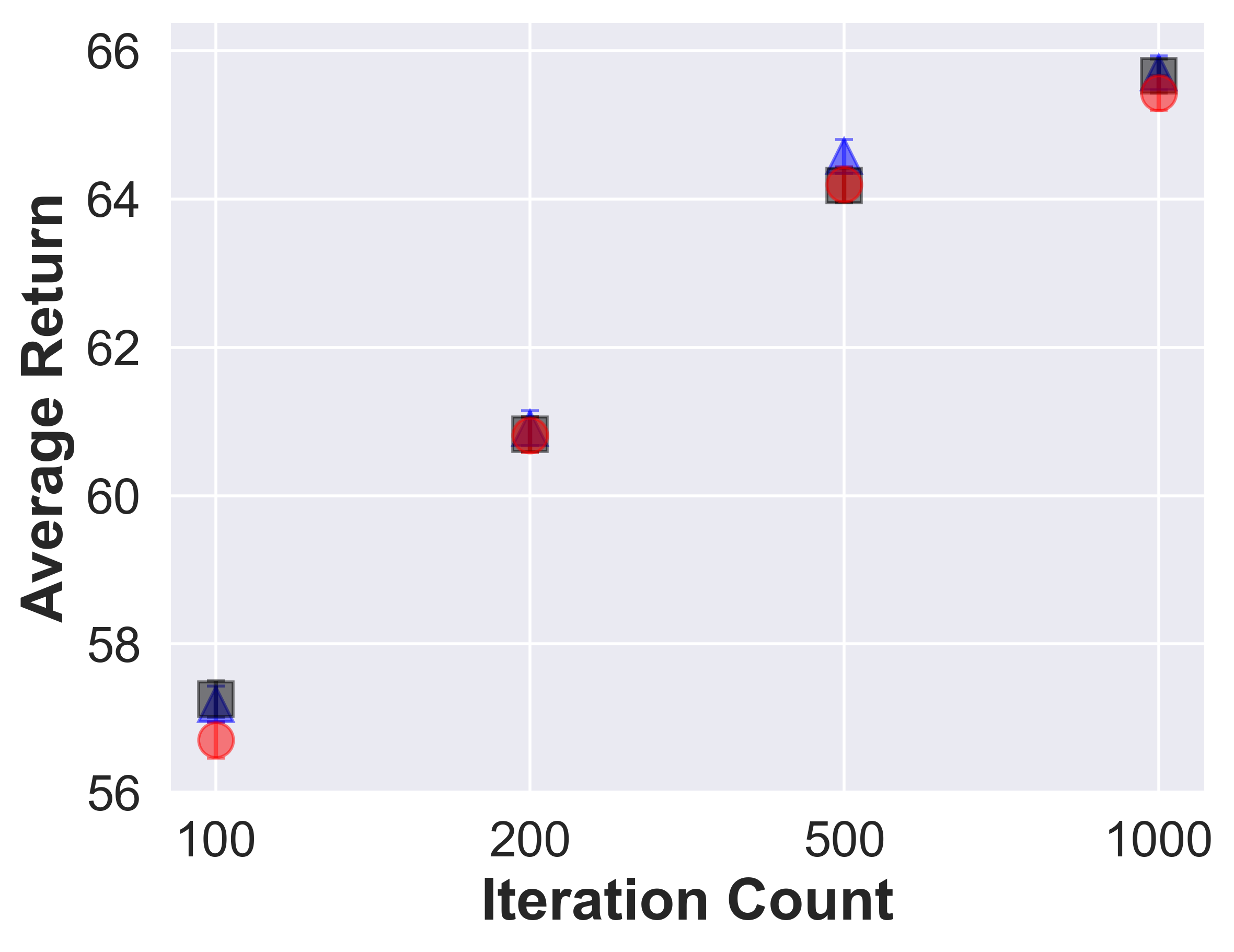}
\caption*{(s) Saving}
\end{minipage}
\hfill
\begin{minipage}{0.23\textwidth}
\centering
\includegraphics[width=\linewidth]{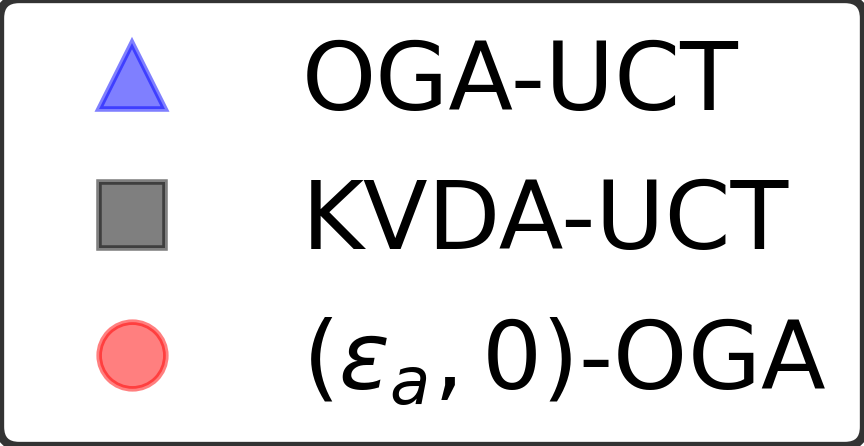}
\caption*{Legend}
\end{minipage}

\caption{The performance plots in dependence of the iteration budget of parameter-optimized KVDA-UCT (our method), OGA-UCT, and $(\varepsilon_{\text{a}},0)$-OGA, $\varepsilon_{\text{a}} > 0$ on all considered environments. Our method KVDA-UCT is either always tightly within the confidence bounds of the top-competitor method or outperforms all competitors simultaneously.}
\label{fig:all_performance_graphs}
\end{figure}

\subsection{Monte Carlo Tree Search}
\label{sec:mcts}

This section specifies the MCTS that was used for the experiments of this paper.

\begin{enumerate}
    \item Our MCTS version builds a directed acyclic graph instead of a tree, i.e. different state-action pairs in the same layer can lead to the same node. This is a necessary condition for ASAP and KVDA to detect any abstractions since they bootstrap of the search graph converging.
    \item We use the Upper Confidence Bounds (UCB) tree policy which is defined as
\begin{equation}
    \text{UCB}(a) = 
    \underbrace{\frac{V_a}{N_a}}_{\text{Q term}} + 
    \underbrace{\lambda \sqrt{\frac{\log\left(\sum\limits_{a^{\prime} \in \mathbb{A}(s)}N_{a^{\prime}}\right)}{N_a}}}_{\text{Exploration term}},
\end{equation}
where, $s$ is the state at which the decision has to be made, $V_a$ is the sum of returns of the action under consideration, and $N_a$ are its visits. The tree policy chooses the action that maximizes the UCB value.
    \item We use the greedy decision policy, which is choosing the root action, after the search statistics have been gathered, with the maximum Q value.
\end{enumerate}

\subsection{Problem descriptions}
\label{sec:problem_descriptions}
The descriptions for all the problem models considered here are either found in \cite{mysurvey} or in \cite{demcts}. Most models are parametrized (e.g. choosing a concrete racetrack for Racetrack), the concrete instance used for the experiments is found in the \textit{ExperimentConfigs} folder in our publicly available implementation \citep{repo}.
Next, we describe the heuristic functions used for the two-player games.

\begin{enumerate}
    \item \textbf{Connect 4}: As the heuristic from the perspective of player one is given by 
    \begin{equation}
        n_2 + 5 n_3 + 25 n_4 
    \end{equation}
    where $n_i$ is the total number of $i$ stones that are in one row/column/diagonal, divided by $i$, but which are not part of a row/column/diagonal of size $i+1$. 
    \item \textbf{Constrictor}: The heuristics function used for player one is the number of grid cells that player one could reach before its opponent. If player one wins, the value $100$ is added to the heuristic.
    \item \textbf{Othello}: The heuristic function for any state for player one is given by a weighted sum of all the occupied grid cells. The weight $w$ of cell $(x,y)$ is given by $10 / (1+d)$ where $d$ is the distance to the closest corner cell. The weight $w$ is positive for any cells occupied by stones of player one and negative for those occupied by player two. In the round player one wins, $100$ is added to the heuristics value and $-100$ if player two won.
    \item \textbf{Pusher}:  We used the heuristic for player one that is equal to the difference of alive units of player one and player two in addition to the Manhatten distance between player one and player two's units' center of mass. In the turn that player one wins, $100$ is added to the heuristic value.
\end{enumerate}

%% file: main.bbl
\begin{thebibliography}{20}
\providecommand{\natexlab}[1]{#1}
\providecommand{\url}[1]{\texttt{#1}}
\expandafter\ifx\csname urlstyle\endcsname\relax
  \providecommand{\doi}[1]{doi: #1}\else
  \providecommand{\doi}{doi: \begingroup \urlstyle{rm}\Url}\fi

\bibitem[Anand et~al.(2015)Anand, Grover, Mausam, and Singla]{AnandGMS15}
Ankit Anand, Aditya Grover, Mausam, and Parag Singla.
\newblock {{ASAP-UCT:} Abstraction of State-Action Pairs in {UCT}}.
\newblock In Qiang Yang and Michael~J. Wooldridge (eds.), \emph{Proceedings of the Twenty-Fourth International Joint Conference on Artificial Intelligence, {IJCAI} 2015, Buenos Aires, Argentina, July 25-31, 2015}, pp.\  1509--1515. {AAAI} Press, 2015.
\newblock URL \url{http://ijcai.org/Abstract/15/216}.

\bibitem[Anand et~al.(2016)Anand, Noothigattu, Mausam, and Singla]{OGAUCT}
Ankit Anand, Ritesh Noothigattu, Mausam, and Parag Singla.
\newblock {OGA-UCT: on-the-go abstractions in UCT}.
\newblock In \emph{Proceedings of the Twenty-Sixth International Conference on International Conference on Automated Planning and Scheduling}, ICAPS'16, pp.\  29–37. AAAI Press, 2016.
\newblock ISBN 1577357574.

\bibitem[Browne et~al.(2012)Browne, Powley, Whitehouse, Lucas, Cowling, Rohlfshagen, Tavener, Liebana, Samothrakis, and Colton]{BrownePWLCRTPSC12}
Cameron Browne, Edward~Jack Powley, Daniel Whitehouse, Simon~M. Lucas, Peter~I. Cowling, Philipp Rohlfshagen, Stephen Tavener, Diego~Perez Liebana, Spyridon Samothrakis, and Simon Colton.
\newblock {A Survey of Monte Carlo Tree Search Methods}.
\newblock \emph{{IEEE} Trans. Comput. Intell. {AI} Games}, 4\penalty0 (1):\penalty0 1--43, 2012.
\newblock \doi{10.1109/TCIAIG.2012.2186810}.
\newblock URL \url{https://doi.org/10.1109/TCIAIG.2012.2186810}.

\bibitem[Grzes et~al.(2014)Grzes, Hoey, and Sanner]{grzes2014ippc}
Marek Grzes, Jesse Hoey, and Scott Sanner.
\newblock {{International Probabilistic Planning Competition (IPPC) 2014}}.
\newblock In \emph{Proceedings of the International Conference on Automated Planning and Scheduling (ICAPS)}, 2014.

\bibitem[Hostetler et~al.(2015)Hostetler, Fern, and Dietterich]{HostetlerFD15}
Jesse Hostetler, Alan Fern, and Thomas~G. Dietterich.
\newblock {Progressive Abstraction Refinement for Sparse Sampling}.
\newblock In Marina Meila and Tom Heskes (eds.), \emph{Proceedings of the Thirty-First Conference on Uncertainty in Artificial Intelligence, {UAI} 2015, July 12-16, 2015, Amsterdam, The Netherlands}, pp.\  365--374. {AUAI} Press, 2015.
\newblock URL \url{http://auai.org/uai2015/proceedings/papers/81.pdf}.

\bibitem[Jiang et~al.(2014)Jiang, Singh, and Lewis]{uctJiang}
Nan Jiang, Satinder Singh, and Richard~L. Lewis.
\newblock {Improving {UCT} planning via approximate homomorphisms}.
\newblock In Ana L.~C. Bazzan, Michael~N. Huhns, Alessio Lomuscio, and Paul Scerri (eds.), \emph{International conference on Autonomous Agents and Multi-Agent Systems, {AAMAS} '14, Paris, France, May 5-9, 2014}, pp.\  1289--1296. {IFAAMAS/ACM}, 2014.
\newblock URL \url{http://dl.acm.org/citation.cfm?id=2617453}.

\bibitem[Kocsis \& Szepesv{\'{a}}ri(2006)Kocsis and Szepesv{\'{a}}ri]{KocsisS06}
Levente Kocsis and Csaba Szepesv{\'{a}}ri.
\newblock {Bandit Based Monte-Carlo Planning}.
\newblock In Johannes F{\"{u}}rnkranz, Tobias Scheffer, and Myra Spiliopoulou (eds.), \emph{Machine Learning: {ECML} 2006, 17th European Conference on Machine Learning, Berlin, Germany, September 18-22, 2006, Proceedings}, volume 4212 of \emph{Lecture Notes in Computer Science}, pp.\  282--293. Springer, 2006.
\newblock \doi{10.1007/11871842\_29}.
\newblock URL \url{https://doi.org/10.1007/11871842\_29}.

\bibitem[Saisubramanian et~al.(2017)Saisubramanian, Zilberstein, and Shenoy]{saisubramanian2017optimizing}
S.~Saisubramanian, S.~Zilberstein, and P.~Shenoy.
\newblock {Optimizing Electric Vehicle Charging Through Determinization}.
\newblock In \emph{ICAPS Workshop on Scheduling and Planning Applications}, 2017.

\bibitem[Schmöcker(2025)]{repo}
Robin Schmöcker.
\newblock {Grouping nodes with known value differences: A lossless UCT-based abstraction algorithm}, 2025.
\newblock Repository available at: \url{https://github.com/codebro634/KVDA_UCT.git}.

\bibitem[Schmöcker \& Dockhorn(2025)Schmöcker and Dockhorn]{mysurvey}
Robin Schmöcker and Alexander Dockhorn.
\newblock A survey of non-learning-based abstractions for sequential decision-making.
\newblock \emph{IEEE Access}, 13:\penalty0 100808--100830, 2025.
\newblock \doi{10.1109/ACCESS.2025.3572830}.

\bibitem[Schmöcker et~al.(2025{\natexlab{a}})Schmöcker, Dockhorn, and Rosenhahn]{aupo}
Robin Schmöcker, Alexander Dockhorn, and Bodo Rosenhahn.
\newblock Aupo - abstracted until proven otherwise: A reward distribution based abstraction algorithm, 2025{\natexlab{a}}.
\newblock URL \url{https://arxiv.org/abs/2510.23214}.

\bibitem[Schmöcker et~al.(2025{\natexlab{b}})Schmöcker, Dockhorn, and Rosenhahn]{intra}
Robin Schmöcker, Alexander Dockhorn, and Bodo Rosenhahn.
\newblock Investigating intra-abstraction policies for non-exact abstraction algorithms, 2025{\natexlab{b}}.
\newblock URL \url{https://arxiv.org/abs/2510.24297}.

\bibitem[Schmöcker et~al.(2025{\natexlab{c}})Schmöcker, Kampmann, and Dockhorn]{ogacad}
Robin Schmöcker, Lennart Kampmann, and Alexander Dockhorn.
\newblock Time-critical and confidence-based abstraction dropping methods.
\newblock In \emph{2025 IEEE Conference on Games (CoG)}, 2025{\natexlab{c}}.
\newblock \doi{10.1109/CoG64752.2025.11114261}.

\bibitem[Schmöcker et~al.(2025{\natexlab{d}})Schmöcker, Schnell, and Dockhorn]{demcts}
Robin Schmöcker, Christoph Schnell, and Alexander Dockhorn.
\newblock Investigating scale independent uct exploration factor strategies, 2025{\natexlab{d}}.
\newblock URL \url{https://arxiv.org/abs/2510.21275}.

\bibitem[Silver et~al.(2017)Silver, Hubert, Schrittwieser, Antonoglou, Lai, Guez, Lanctot, Sifre, Kumaran, Graepel, Lillicrap, Simonyan, and Hassabis]{alphazero}
David Silver, Thomas Hubert, Julian Schrittwieser, Ioannis Antonoglou, Matthew Lai, Arthur Guez, Marc Lanctot, Laurent Sifre, Dharshan Kumaran, Thore Graepel, Timothy~P. Lillicrap, Karen Simonyan, and Demis Hassabis.
\newblock {Mastering Chess and Shogi by Self-Play with a General Reinforcement Learning Algorithm}.
\newblock \emph{CoRR}, abs/1712.01815, 2017.
\newblock URL \url{http://arxiv.org/abs/1712.01815}.

\bibitem[Sokota et~al.(2021)Sokota, Ho, Ahmad, and Kolter]{SokotaHAK21}
Samuel Sokota, Caleb Ho, Zaheen~Farraz Ahmad, and J.~Zico Kolter.
\newblock {Monte Carlo Tree Search With Iteratively Refining State Abstractions}.
\newblock In Marc'Aurelio Ranzato, Alina Beygelzimer, Yann~N. Dauphin, Percy Liang, and Jennifer~Wortman Vaughan (eds.), \emph{Advances in Neural Information Processing Systems 34: Annual Conference on Neural Information Processing Systems 2021, NeurIPS 2021, December 6-14, 2021, virtual}, pp.\  18698--18709, 2021.
\newblock URL \url{https://proceedings.neurips.cc/paper/2021/hash/9b0ead00a217ea2c12e06a72eec4923f-Abstract.html}.

\bibitem[Sutton \& Barto(2018)Sutton and Barto]{sutton2018reinforcement}
Richard~S. Sutton and Andrew~G. Barto.
\newblock \emph{{Reinforcement Learning: An Introduction}}.
\newblock The MIT Press, 2nd edition, 2018.

\bibitem[Xu et~al.(2023)Xu, Dockhorn, and Perez-Liebana]{EMCTSXu}
Linjie Xu, Alexander Dockhorn, and Diego Perez-Liebana.
\newblock {Elastic Monte Carlo Tree Search}.
\newblock \emph{IEEE Transactions on Games}, 15\penalty0 (4):\penalty0 527--537, 2023.
\newblock \doi{10.1109/TG.2023.3282351}.

\bibitem[Yoon et~al.(2007)Yoon, Fern, and Givan]{YoonFG07}
Sung~Wook Yoon, Alan Fern, and Robert Givan.
\newblock {FF-Replan: {A} Baseline for Probabilistic Planning}.
\newblock In Mark~S. Boddy, Maria Fox, and Sylvie Thi{\'{e}}baux (eds.), \emph{Proceedings of the Seventeenth International Conference on Automated Planning and Scheduling, {ICAPS} 2007, Providence, Rhode Island, USA, September 22-26, 2007}, pp.\  352. {AAAI}, 2007.
\newblock URL \url{http://www.aaai.org/Library/ICAPS/2007/icaps07-045.php}.

\bibitem[Yoon et~al.(2008)Yoon, Fern, Givan, and Kambhampati]{YoonFGK08}
Sung~Wook Yoon, Alan Fern, Robert Givan, and Subbarao Kambhampati.
\newblock {Probabilistic Planning via Determinization in Hindsight}.
\newblock In Dieter Fox and Carla~P. Gomes (eds.), \emph{Proceedings of the Twenty-Third {AAAI} Conference on Artificial Intelligence, {AAAI} 2008, Chicago, Illinois, USA, July 13-17, 2008}, pp.\  1010--1016. {AAAI} Press, 2008.
\newblock URL \url{http://www.aaai.org/Library/AAAI/2008/aaai08-160.php}.

\end{thebibliography}
